\documentclass[journal]{IEEEtran}
\usepackage[nocompress]{cite}
\usepackage{algorithmic}
\usepackage{array}
\usepackage{subfigure}
\usepackage{fixltx2e}
\usepackage{stfloats}
\usepackage{url}
\usepackage{color}
\usepackage[pagebackref=true,breaklinks=true,letterpaper=true,colorlinks,bookmarks=false]{hyperref}
\usepackage{caption}

\usepackage{epsfig}
\usepackage{graphicx}
\usepackage{amsmath}
\usepackage{amssymb}

\usepackage{array}
\usepackage{multirow}
\usepackage{soul}
\usepackage{siunitx}

\usepackage{colortbl}
\usepackage{color}
\usepackage{xcolor}
\usepackage{tabularx}

\usepackage{xspace}
\makeatletter
\DeclareRobustCommand\onedot{\futurelet\@let@token\@onedot}
\def\@onedot{\ifx\@let@token.\else.\null\fi\xspace}

\def\eg{\emph{e.g}\onedot} 
\def\ie{\emph{i.e}\onedot}

\def\etal{\emph{et al}\onedot}
\makeatother

\newcommand{\PreserveBackslash}[1]{\let\temp=\\#1\let\\=\temp}
\newcolumntype{C}[1]{>{\PreserveBackslash\centering}p{#1}}

\newcommand{\para}[1]{\par\noindent\textbf{#1}}
\newcommand{\comm}[1]{{\small \textcolor{red}{\emph{}}}}
\newcommand{\redb}[1]{\textbf{\textcolor{red}{#1}}}

\newcommand{\blank}[1]{#1}

\graphicspath{{images/}{biography/}}

\begin{document}

\title{Learning Spatial Attention for Face Super-Resolution}

\author{Chaofeng Chen, Dihong Gong, Hao Wang, Zhifeng Li,~\IEEEmembership{Senior,~IEEE},
        and Kwan-Yee~K.~Wong,~\IEEEmembership{Senior,~IEEE}
\IEEEcompsocitemizethanks{\IEEEcompsocthanksitem Chaofeng Chen and Kwan-Yee~K.~Wong are with the Department of Computer Science, The University of Hong Kong, Pokfulam Road, Hong Kong. (E-mail: [cfchen, kykwong]@cs.hku.hk)
\IEEEcompsocthanksitem Dihong Gong, Hao Wang and Zhifeng Li are with Tencent AI Lab, Shenzhen 518000, China. (
E-mail: gongdihong@gmail.com, hawelwang@tencent.com, michaelzfli@tencent.com)}
}

\markboth{IEEE Transactions on Image Processing}%
{Shell \MakeLowercase{\textit{et al.}}: Bare Demo of IEEEtran.cls for Computer Society Journals}

\maketitle

\begin{abstract}

General image super-resolution techniques have difficulties in recovering detailed face structures when applying to low resolution face images. Recent deep learning based methods tailored for face images have achieved improved performance by jointly trained with additional task such as face parsing and landmark prediction. However, multi-task learning requires extra manually labeled data. Besides, most of the existing works can only generate relatively low resolution face images (\eg, $128\times128$), and their applications are therefore limited. In this paper, we introduce a novel SPatial Attention Residual Network (SPARNet) built on our newly proposed Face Attention Units (FAUs) for face super-resolution. Specifically, we introduce a spatial attention mechanism to the vanilla residual blocks. This enables the convolutional layers to adaptively bootstrap features related to the key face structures and pay less attention to those less feature-rich regions. This makes the training more effective and efficient as the key face structures only account for a very small portion of the face image. Visualization of the attention maps shows that our spatial attention network can capture the key face structures well even for very low resolution faces (\eg, $16\times16$). Quantitative comparisons on various kinds of metrics (including PSNR, SSIM, identity similarity, and landmark detection) demonstrate the superiority of our method over current state-of-the-arts. We further extend SPARNet with multi-scale discriminators, named as SPARNetHD, to produce high resolution results (\ie, $512\times512$). We show that SPARNetHD trained with synthetic data cannot only produce high quality and high resolution outputs for synthetically degraded face images, but also show good generalization ability to real world low quality face images. Codes are available at \url{https://github.com/chaofengc/Face-SPARNet}.
\end{abstract}

\begin{IEEEkeywords}
  Face Super-Resolution, Spatial Attention, Generative Adversarial Networks
\end{IEEEkeywords}

%
\IEEEpeerreviewmaketitle

\section{Introduction}\label{sec:introduction}

Face super-resolution (SR), also known as face hallucination, refers to generating high resolution (HR) face images from the corresponding low resolution (LR) inputs. Since there exist many low resolution face images (\eg, faces in surveillance videos ) and face analysis algorithms (\eg, face recognition) often perform poorly on such images, there is a growing interest in face SR.

Different from general image SR, face SR places its emphasis on the recovery of \blank{the} key \blank{face} structures (\ie, shapes of \blank{face} components and face outline). These structures only account for a very small portion of the image, but are often more difficult to recover as they exhibit larger pixel variations. Training a deep neural network with the commonly used mean square error (MSE) loss, which weights pixels equally, is not very effective in recovering these ``sparse'' structures. \blank{Previous} works \cite{Chen2018CVPR,bulat2017super,yu2018face} proposed incorporating additional task, such as face parsing and landmark detection, to assist the training of face SR networks. \cite{Chen2018CVPR,yu2018face} also used predicted face priors to help face SR. Although joint training with these additional tasks helps enhance the importance of \blank{the} key \blank{face} structures, there are two major drawbacks\blank{, namely} (1) extra effort is needed to label the data for the additional \blank{task,} and (2) predicting face prior from LR inputs is \blank{itself also} a difficult problem.

On the other hand, if we subdivide a face image into many small regions and consider each region as an individual sample, the unbalanced distribution between regions containing key \blank{face} structures (referred to as hard regions) and those not containing key \blank{face} structures (referred to as easy regions) will \blank{resemble} the imbalance between foreground and background samples in object detection. This suggests that we may adopt some techniques similar to {\em bootstrapping} or {\em online hard example mining} (OHEM)~\cite{shrivastavaCVPR16ohem} in object detection to solve our face SR problem.

\begin{figure}[t]
  \includegraphics[width=.99\linewidth]{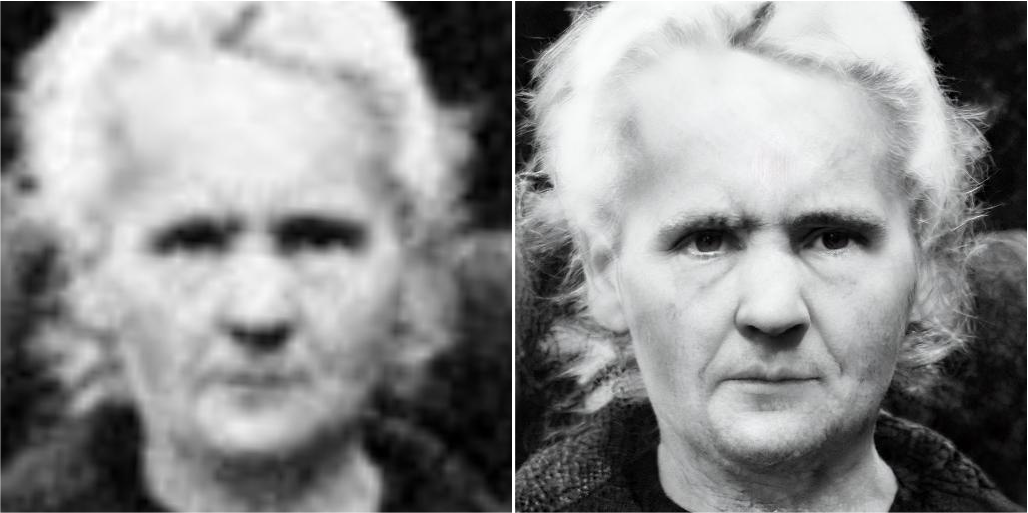} 
  \caption{\blank{Super-resolution} result \blank{produced by} SPARNetHD for an old photo of Marie Curie from Solvay conference 1927. Please zoom in to see details.}
  \label{fig:intro_1}
\end{figure}  

In this paper, we introduce a carefully designed \blank{Face Attention Unit (FAU)} to construct a \blank{SPatial Attention Residual Network} (SPARNet) for face SR. The key idea is to bootstrap features related to the key \blank{face} structures \blank{using} a 2D spatial attention map. Instead of hard selection, the spatial attention map assigns a score between $0$ and $1$ to each spatial location of the feature map. This allows learning \blank{the prediction of} the spatial attention map through gradient descent. Spatial attention maps in different FAUs of the network \blank{can learn} to focus on different \blank{face} structures. For example, attention maps in \blank{deeper layers} focus more on coarse structures such as eyes and mouth, while those in \blank{shallower layers} focus more on detailed textures such as hairs. Considering that most \blank{of the} existing face SR methods can only \blank{produce} $128\times128$ outputs, we further extend SPARNet, referred to as SPARNetHD, to generate \blank{high resolution outputs (\ie, $512\times512$)}. Specifically, we enlarge the output resolution of SPARNet from \blank{$128\times128$ to $512\times512$} and adopt \blank{a} multi-scale discriminator loss similar to Pix2PixHD~\cite{wang2018pix2pixHD} to generate more realistic textures. Experiments show that SPARNetHD trained with synthetic LR data is pretty robust with natural LR inputs, while models without the proposed spatial attention mechanism produce undesirable artifacts. We show an example result of SPARNetHD on an old photo in \blank{Fig.~\ref{fig:intro_1}}. We can see that \blank{SPARNetHD} can restore key \blank{face} components very well and also generate high resolution and realistic textures. 

\blank{The} key contributions of this paper can be summarized as follows:

\begin{enumerate}
  \item We propose an efficient framework \blank{named} SPARNet for face super-resolution. Without relying on any extra supervisions (\eg, face parsing maps and landmarks), it achieves state-of-the-art performance on various kinds of metrics, including PSNR, SSIM, identity similarity, and landmark detection.
  \item We show that the proposed FAU\blank{, the basic building block of SPARNet,} can bootstrap the key \blank{face} structures (\ie, \blank{face} components and face outline) and significantly improve the performance of face super-resolution.
  \item By repeating FAUs in SPARNet, the spatial attention maps in different FAUs can learn to focus on different \blank{face} structures and further \blank{improve} the performance of SPARNet.
  \item We introduce SPARNetHD to \blank{generate} high resolution \blank{face} images (\ie, $512\times512$), and the model trained with synthetic data works \blank{equally} well on natural LR images.   
\end{enumerate}

\begin{figure*}[t]
  \centering
  \captionsetup{justification=centering}
  \includegraphics[width=1.\linewidth]{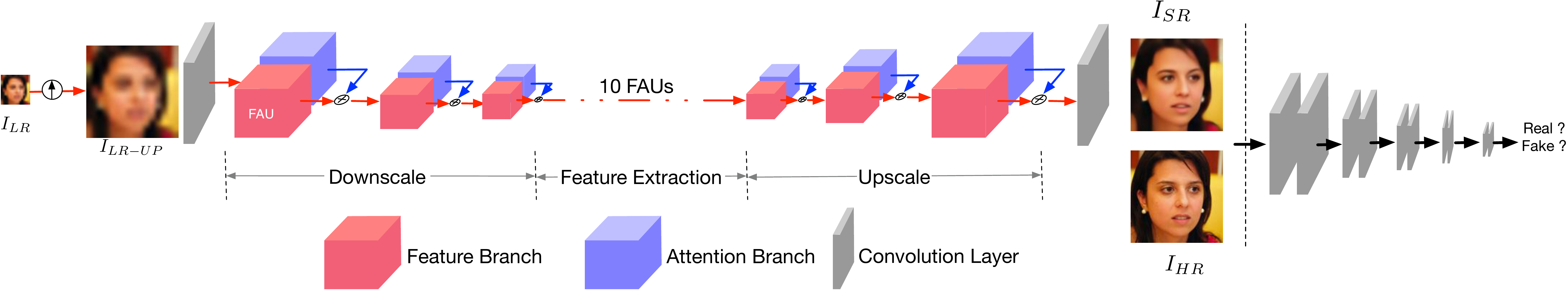}
  \caption{Architecture of the proposed SPatial Attention Residual Network (SPARNet).}
  \label{fig:arch}
\end{figure*}

\section{Related Works}

In this section, we briefly review the literature on face super-resolution and recent attention neural networks.

\subsection{Face Super-Resolution}
Face hallucination was pioneered by Baker and Kanade \cite{baker2002limits}, who showed that it is possible to perform high magnification SR for images of a specific category (\eg, face and text). Since then many methods were proposed to improve the performance of face hallucination. They can be roughly classified into sub-space based methods \cite{wang2005hallucinating,liu2007face,ma2010hallucinating,liu2001two,liu2005hallucinating} and component based methods \cite{tappen2012bayesian,yang2013structured,song-ijcai17-faceSR}. Sub-space based methods usually \blank{rely} on Principle Component Analysis (PCA), which requires precisely aligned faces. Component based methods require detecting facial landmarks, which is difficult for LR face images. Both kinds of methods fail to produce satisfactory results for face SR with a high upscale factor.

Recently, deep convolutional neural networks (CNNs) have brought remarkable progress to face SR. Zhu \etal \cite{zhu2016deep} proposed a cascaded two-branch network to optimize face hallucination and dense correspondence field estimation in a unified framework. Yu \etal exploited generative adversarial networks (GAN) \cite{goodfellow2014generative} to directly super-resolve the LR inputs. They further improved their model to handle unaligned faces \cite{yu2017face}, noisy faces \cite{yu2017hallucinating}, and faces with different attributes \cite{yusuper}. Instead of directly inferring HR face images, Huang \etal proposed to predict wavelet coefficients from LR images to reconstruct HR images. Latest works employed extra \blank{face} prior supervisions, such as face parsing maps \cite{Chen2018CVPR}, landmark heatmaps \cite{bulat2017super,yu2018face}, and identity information \cite{zhang2018super}, to train their networks. Kim \etal \cite{pfsrnet} proposed a facial attention loss which focuses the network on the landmark region. Ma \etal \cite{dicnet} introduced an iterative method which predicts SR results and landmarks iteratively. Although extra supervisions help to improve the performance, there are two shortcomings\blank{, namely} (1) extra effort is needed to label the data, and (2) it is \blank{an} indirect way to \blank{direct} the SR network \blank{towards the key face} structures. Different from these methods, our SPARNet learns \blank{to predict} spatial attention maps to bootstrap key \blank{face} structures and \blank{other} feature-rich regions. 

\subsection{Attention Networks}
Attention mechanism has been widely used in high level vision tasks, such as image classification \cite{mnih2014recurrent,hu2018senet,wang2017residual,woo2018cbam}, image captioning \cite{xu2015show,chen2017sca}\blank{,} and visual question answering \cite{schwartz2017high,yu2017multi}. The key idea is to \blank{reweight} features using a score map to emphasize \blank{important features and suppress less useful ones} \cite{hu2018senet}. Wang \etal \cite{wang2017residual} introduced a trunk-and-mask attention mechanism to a residual network for image classification. He \etal \cite{hu2018senet} proposed the Squeeze-and-Excitation network which
\blank{employs} a channel-wise attention mechanism and \blank{demonstrates} significant performance improvements. Woo \etal proposed a convolutional block attention module (CBAM) which sequentially infers attention maps along the channel and spatial dimensions separately. Attention mechanism has also been employed in image generation tasks recently. Zhang \etal \cite{zhang2018rcan} combined channel attention with a very deep residual network for image SR. Cao \etal \cite{Cao2017Attention} proposed an attention aware face SR framework based on reinforcement learning, which sequentially attends to, crops out\blank{,} and super-resolves a patch.

Although closely related, our work is different from \cite{zhang2018rcan} and \cite{Cao2017Attention} in several aspects. First, attention mechanisms in high level tasks \blank{often} employ pooling for extracting semantic information. For face SR, however, this may \blank{result the loss of} important low and middle level features such as edges and shapes. In contrast, our attention mechanism \blank{is designed to take} advantages of multi-scale features. Second, \blank{unlike \cite{zhang2018rcan} which utilizes channel attention, our work considers spatial attention which facilitates region based attention of the key face components.} 
Third, \blank{unlike \cite{Cao2017Attention} which takes a patch based approach, we generate spatial attention maps for the entire face.} This allows our network to have a global view of the \blank{face} structures and benefit from the contextual information.

\subsection{High Resolution Image Generation with GAN.} Recently, GAN has demonstrated to be very effective in generating HR images. Wang \etal \cite{wang2018pix2pixHD} proposed Pix2PixHD and used a multi-scale generator and discriminator architecture for HR image generation. Karras \etal \cite{karras2017progressive,karras2019style} proposed a new progressive training methodology for GAN to generate HR face images from random input vectors, and improved the quality of synthesized faces in \cite{karras2019analyzing}. Park \etal \cite{park2019SPADE} proposed the spatially-adaptive normalization layer for generating photo-realistic images given an input semantic layout. We follow the idea of Pix2PixHD and use \blank{a} multi-scale discriminator to generate \blank{high resolution} results. Different from Pix2PixHD, our SPARNetHD does not require progressive training from LR to HR images\blank{,} and our results are much better than Pix2PixHD with the help of the proposed spatial attention mechanism.    

\section{SPARNet for Face SR}

\subsection{Overview} \label{sec:overview}
As shown in Fig.~\ref{fig:arch}, our SPARNet consists of three modules, namely the downscale module, the feature extraction module\blank{,} and the upscale module. Each of these modules is composed of a stack of FAUs (see \ref{sec:FAU} for details). Let 
$I_{LR} \in \mathbb{R}^{3\times H'\times W'}$, $I_{SR} \in \mathbb{R}^{3\times H\times W}$\blank{,} and $I_{HR} \in \mathbb{R}^{3\times H\times W}$ denote the LR face image (\ie, input), the super-resolved image (\ie, output)\blank{,} and the ground truth HR image respectively. $I_{LR}$ is first upsampled to the same spatial dimension as $I_{HR}$ through bicubic interpolation, denoted as $I_{LR-UP} \in \mathbb{R}^{3\times H\times W}$, which is then fed to SPARNet to produce $I_{SR}$. Given a training set with $N$ pairs of LR-HR images, $\left \{ I_{LR}^i, I_{HR}^i\right\}_{i=1}^N$, we optimize SPARNet by minimizing the pixel-level $L_2$ loss given by 
\begin{equation}
    \mathcal{L}_{pix}(\Theta) = \frac{1}{N} \blank{\sum_{i=1}^N} \|\mathcal{F}_{SPAR}(I_{LR-UP}^i, \Theta) - I_{HR}^i\blank{\|_2^2}, \label{eq:pix}
\end{equation}
where $\mathcal{F}_{SPAR}$ and $\Theta$ denote \blank{SPARNet and its network parameters} respectively.

\subsection{Face Attention Unit}\label{sec:FAU}
Based on the observation that some \blank{face} parts (\eg, key \blank{face} components like eyes, eyebrows, nose, and mouth) are more important than the others (\eg, shading of the cheek) in face SR, we propose a spatial attention mechanism to make our network focus more on the important and informative features. The key question is how to produce the attention map and how to integrate it with the convolutional layers. First, \blank{we believe} the spatial attention mechanism should not only have a high level view of the face but should also focus on low level structures. \blank{Note that a} high level view helps the network to learn how faces look, while a low level view makes the network \blank{learn local details better.} \blank{Hence, it would be desirable for the spatial attention mechanism to be able to} learn from multi-scale features. Second, residual blocks have \blank{demonstrated} great success in both general SR task \cite{ledig2017photo,Lim_2017_CVPR_Workshops,zhang2018rcan} and face SR task \cite{Chen2018CVPR,bulat2017super}. It should \blank{therefore} be beneficial to integrate spatial attention mechanism with residual blocks.

\begin{figure}[t]
  \centering
  \includegraphics[width=.9\linewidth]{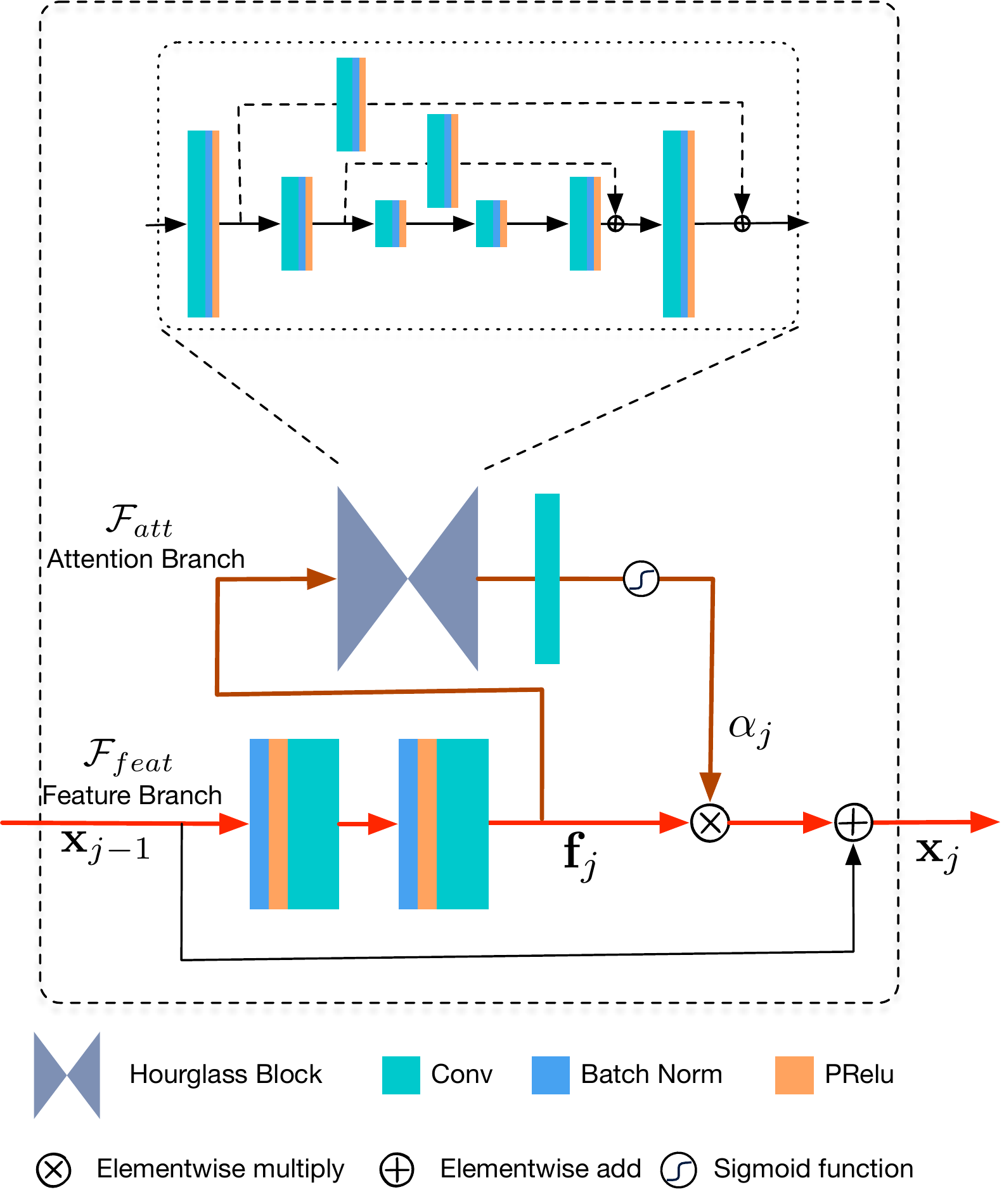}
  \caption{Face Attention Unit}
  \label{fig:sparb}
\end{figure}

Based on the above \blank{discussion}, we propose a Face Attention Unit (FAU) which extends the original residual block by introducing a spatial attention branch (see Fig.~\ref{fig:sparb}). By stacking FAUs together, important features for face SR are continuously enhanced. Denote the feature input \blank{of the $j$-th indexed FAU} as $\blank{\mathbf{x}_{j-1}} \in \mathbb{R}^{C_{j-1}\times H_{j-1} \times W_{j-1}}$, the attention map $\blank{{\boldsymbol \alpha}_{j}} \in \mathbb{R}^{1\times H_j \times W_j}$ is computed as
\begin{align}
    \blank{\mathbf{f}_j} &= \mathcal{F}_{feat}(\blank{\mathbf{x}_{j-1}}), \\
    \blank{\boldsymbol{\alpha}_j} &= \sigma(\mathcal{F}_{att}(\blank{\mathbf{f}_j})),
\end{align}
where $\blank{\mathbf{f}_j} \in \mathbb{R}^{C_{j}\times H_{j} \times W_{j}}$ is the output of the feature branch $\mathcal{F}_{feat}$, $\mathcal{F}_{att}$ \blank{denotes} the attention branch, and $\sigma$ is the sigmoid function. Finally, the output of \blank{the $j$-th indexed} FAU is \blank{given by}
\begin{equation}
    \blank{\mathbf{x}_j = \mathbf{x}_{j-1} + {\boldsymbol \alpha}_j \otimes \mathbf{f}_j}, \label{eq:res-out}
\end{equation}
where ``$\otimes$'' denotes element-wise multiplication. Details of \blank{$\mathcal{F}_{feat}$ and $\mathcal{F}_{att}$} are \blank{given in the next two paragraphs.}

\para{Attention Branch} As discussed above, the attention branch should extract multi-scale features. We adopt the hourglass block followed by an extra Conv layer to generate the attention map. The hourglass block is known to be capable of capturing information at multiple scales \cite{newell2016stacked}. It has also shown great performance in face \blank{analysis} tasks, such as face alignment~\cite{bulat2017far} and face parsing~\cite{Chen2018CVPR}. The kernel size and filter number for all Conv layers in the hourglass block are $3\times3$ and $64$ respectively.

\begin{figure}[htb]
  \includegraphics[width=1.\linewidth]{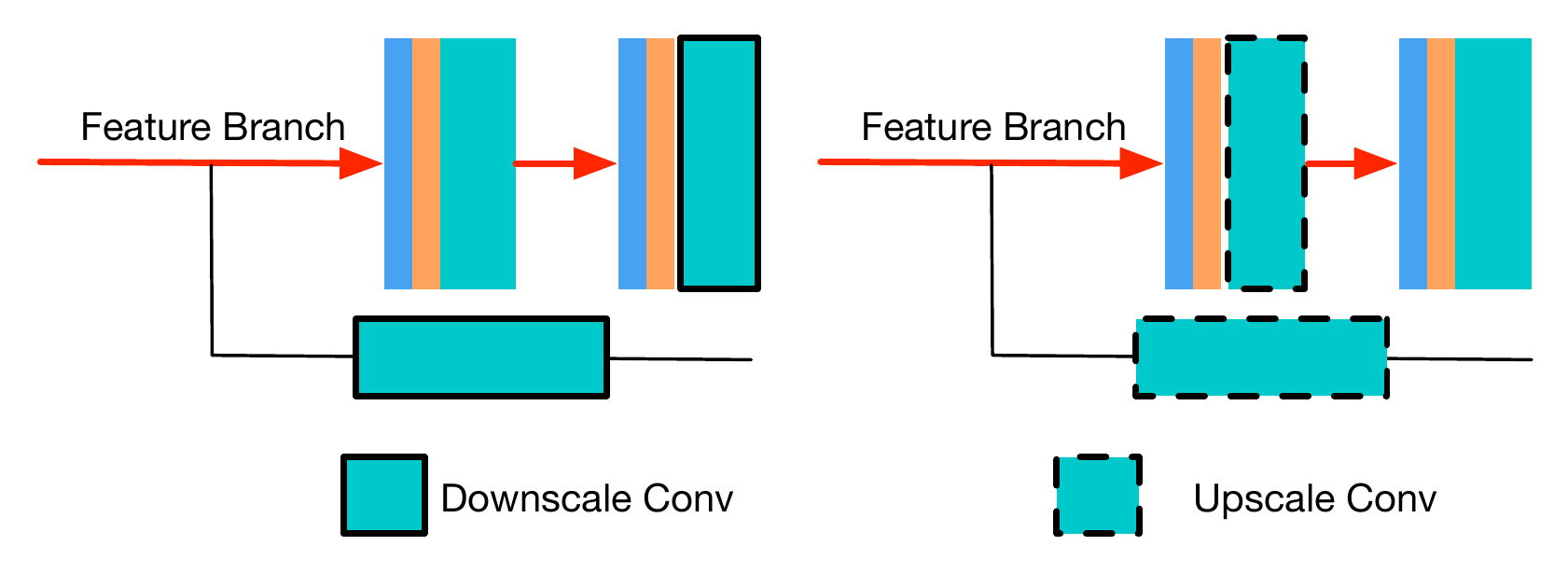}
  \caption{Scale Residual Block}
  \label{fig:scale_res}
\end{figure}

\para{Feature Branch} After experimenting with several variants of residual blocks \cite{Lim_2017_CVPR_Workshops,He2016}, we finally choose the pre-activation Residual Unit with PReLU \cite{he2015delving} as our feature branch. Although previous general image SR work \cite{Lim_2017_CVPR_Workshops} argued that networks without batch normalization perform better, we find that pre-activation structure shows slightly better performance with batch normalization. We use PReLU as the activation function after batch normalization to avoid ``dead features'' caused by zero gradients in ReLU, and it shows more stable performance \cite{dong2016accelerating}.
For the residual blocks in downscale and upscale module, we slightly modify the original residual branch using scale Conv (see Fig.~\ref{fig:scale_res}), and (\ref{eq:res-out}) becomes
\begin{equation}
    \blank{\mathbf{x}_j = \mathcal{F}_{scale}(\mathbf{x}_{j-1}) + {\boldsymbol \alpha}_j \otimes \mathbf{f}_j},
\end{equation}
where $\mathcal{F}_{scale}$ denotes the scale Conv layer. Downscale Conv is a normal Conv layer with stride $2$, and upscale Conv is a nearest-neighbor upsampling layer with a normal Conv layer which helps avoid checkerboard artifacts \cite{odena2016deconvolution}.

\subsection{Multi-scale Discriminator Network}

We extend \blank{our} SPARNet to SPARNetHD \blank{to generate high resolution} and more realistic SR images. SPARNetHD increases the channel number of SPARNet and adopt multi-scale discriminators similar to Pix2PixHD \cite{wang2018pix2pixHD}. We refer to the discriminators as $D_1, D_2$\blank{,} and $D_3$, which are used to discriminate \blank{SR images at three different scales, namely} $512\times512$, $256\times256$\blank{,} and $128\times128$\blank{, respectively}, from the \blank{ground truth downsampled to the same resolutions}. Using multiple discriminators at different scales can help to improve the quality of the \blank{SR} images. 

The loss \blank{functions used in training} SPARNetHD \blank{are composed of the following} four components: 

\noindent (1) \textit{Pixel loss}. 
\blank{We} use L1-norm as the pixel level loss between $I_{SR}$ and $I_{HR}$. It mainly helps to constrain the low level information in the outputs especially color, and is defined as  
\begin{align}
  \blank{\mathcal{L}_{pix}^{h}} &\blank{= \frac{1}{N} \sum_{i=1}^N \|I_{SR}^i - I_{HR}^i\|_1},\\
  \blank{I_{SR}^i} &\blank{= G(I_{LR-UP}^i)},
\end{align}
\blank{where $G$ denotes the SPARNetHD generator.} 

\noindent (2) \textit{Adversarial loss}. 
\blank{It is the critical loss which helps to make the outputs sharper and generate more realistic textures such as hair. The loss functions of the generator and discriminator are formulated as}
\begin{gather}
  \mathcal{L}^h_{GAN\_G} = \frac{1}{N} \sum_{i=1}^N\sum_{k=1}^3 -D_k(I_{SR}^i),\\
  \begin{aligned}
  \mathcal{L}^h_{GAN\_D} = \frac{1}{N} \sum_{i=1}^N\sum_{k=1}^3 [\max(0, 1 - D_k(I_{HR}^i)) \\ + \max(0, 1 + D_k(I_{SR}^i))],
  \end{aligned}
\end{gather}
where the outputs of $D_k$ are scalars which indicates whether the input images are real ($\geq 1$) or fake ($\leq -1$).

\noindent (3) \textit{Feature matching loss}. This is the feature space loss of the discriminators \cite{wang2018pix2pixHD}. It helps to stabilize the training of GAN. Let \blank{$\mathbf{f}_{D_k}^l$} be \blank{the} feature map of the \blank{$l$-th} layer in \blank{$D_k$}, \blank{$L_k$} be the total number of layers in \blank{$D_k$}, and \blank{$M_k^l$} be the number of elements in \blank{$\mathbf{f}_{D_k}^l$}. \blank{The feature matching loss is then formulated as}
\begin{equation}
  \mathcal{L}^h_{fm} = \blank{\frac{1}{N} \sum_{i=1}^N\sum_{k=1}^3 \sum_{l=1}^{L_k} \frac{1}{M_k^l} \| \mathbf{f}_{D_k}^l(I_{SR}^i) - \mathbf{f}_{D_k}^l(I_{HR}^i)\|_1},\label{eq:fm_loss}
\end{equation}   

\noindent (4) \textit{Perceptual loss}. Different from feature matching loss, perceptual loss~\cite{ledig2017photo} is the feature space loss of the pretrained VGG19 network~\cite{simonyan2014very}. It helps to constrain the high level semantics in the outputs. We follow the notation of Eq.~\ref{eq:fm_loss} and denote the perceptual loss as
\begin{equation}
\mathcal{L}^h_{pcp}=\blank{\frac{1}{N} \sum_{i=1}^N\sum_{l=1}^{L_{VGG}} \frac{1}{M_{VGG}^l} \|\mathbf{f}_{VGG}^l(I_{SR}^i) - \mathbf{f}_{VGG}^l(I_{HR}^i)\|_1.} 
\end{equation}

\noindent \blank{Finally, the loss functions are defined as}
\begin{gather}
  \mathcal{L}^h_G = \lambda_{pix}\mathcal{L}^h_{pix} + \lambda_{adv}\mathcal{L}^h_{GAN\_G} + \lambda_{fm}\mathcal{L}^h_{fm} + \lambda_{pcp}\mathcal{L}^h_{pcp}\blank{,} \label{eq:loss-hd-final}\\
  \mathcal{L}^h_D = \mathcal{L}^h_{GAN\_D}\blank{,} 
\end{gather}
where $\mathcal{L}^h_G$ and $\mathcal{L}^h_D$ are minimized iteratively to train $G$ and $D$, and $\lambda_{pix}, \lambda_{adv}, \lambda_{fm}$, and $\lambda_{pcp}$ are the \blank{weights for each loss item respectively}.

The reason why we use more complicated loss functions to train SPARNetHD is to generate results with high perceptual quality. As pointed out by \cite{blau2018perception}, distortion and perceptual quality are at odds with each other. Methods trained with only $\mathcal{L}_{pix}$ in Eq. \ref{eq:pix} always lead to better PSNR and SSIM scores, but over-smoothed results with bad perceptual quality. Therefore, we train SPARNet with $\mathcal{L}_{pix}$ to make fair comparison on distortion metrics (\ie, PSNR and SSIM scores) with previous works. 

\subsection{Training Details}
For SPARNet, we set the batch size as 64, and fix the learning rate at $2\times10^{-4}$. We \blank{use} Adam~\cite{kingma2014adam} to optimize the model with $\beta_1=0.9$ and $\beta_2=0.99$. For SPARNetHD, we empirically set $\lambda_{pix}=100$, \blank{$\lambda_{adv}=1$}, $\lambda_{fm}=10$, and \blank{$\lambda_{pcp}=1$}. The learning rates of $G$ and $D$ are $1\times10^{-4}$ and $4\times10^{-4}$ respectively. We \blank{use} Adam \blank{to optimize} both $G$ and $D$ with $\beta_1=0.5$ and $\beta_2=0.99$. The batch size \blank{is} set to $2$. Our models \blank{are} implemented in PyTorch and run on a Tesla K40 GPU.

\section{Experiments}

In this section, we conduct experiments for SPARNet and SPARNetHD respectively. First, we \blank{analyze}  the effectiveness of the proposed spatial attention mechanism in SPARNet, and compare SPARNet with previous face SR methods and general SR methods with different evaluation metrics using the same training data. Second, we evaluate the performance of SPARNetHD on real LR faces trained on synthetic datasets. 

\subsection{Analysis of SPARNet}

\subsubsection{Datasets and Evaluation Metrics}

\para{Training Data} We use CelebA~\cite{liu2015faceattributes} to train SPARNet. We first detect faces using MTCNN~\cite{zhang2016mtcnn}, and crop out the face regions roughly without any pre-alignment operation. Next, we select images larger than $128\times128$, resize them to $128\times128$ through bicubic interpolation, and use them as the HR training set. \blank{The} LR training set is obtained by downsampling the HR images to $16\times16$. This results in roughly $179$K image pairs. To avoid overfitting, we \blank{carry} out data augmentation by random horizontal flipping, image rescaling (between $1.0$ and $1.3$), and image rotation ($\ang{90}$, $\ang{180}$ and $\ang{270}$). 

\para{Testing Data} Following previous works~\cite{zhu2016deep,huang2017wavelet,Chen2018CVPR}, we use the test set of Helen~\cite{Sagonas_2013_CVPR_Workshops} for evaluation of image quality and landmark detection. For identity similarity evaluation, we use the same test set as in \cite{zhang2018super}, which is specifically designed to preserve identity in face SR. This dataset contains randomly selected images of $1,000$ identities from UMD-Face~\cite{bansal2016umdfaces}. 

Following the practice of image super-resolution, the Peak Signal-to-Noise Ratio (PSNR) and Structure SIMilarity (SSIM) index calculated on the luminance channel are used as the primary quantitative evaluation metrics. Since two most important applications of face SR are alignment and identification of LR faces, we adopt two further metrics, namely landmark detection and identity similarity, to evaluate face SR performance.

\para{Landmark Detection} As mentioned before, structure recovery is very important in face SR. Similar to \cite{Chen2018CVPR,bulat2017super}, we use landmark detection accuracy for evaluation. Specifically, we use a popular landmark detection model, FAN\footnote{\url{https://github.com/1adrianb/face-alignment}} \cite{bulat2017far}, which is pre-trained on HR faces, to detect landmarks on SR and HR faces. Following \cite{bulat2017far}, we use the \blank{area under the curve} (AUC) of the normalized mean error (NME) to quantify the performance.

\para{Identity Similarity} Identity similarity measures how well identity information is preserved in a super-resolved face. Same as \cite{zhang2018super}, we first extract identity feature vectors for the SR and HR faces using a pre-trained SphereFace model\footnote{\url{https://github.com/clcarwin/sphereface_pytorch}} \cite{Liu_2017_CVPR}, and then compute the identity similarity as the cosine distance between the two feature vectors. 

\begin{figure}[t]
  \centering
  \subfigure[Image Quality]{\includegraphics[width=.49\linewidth]{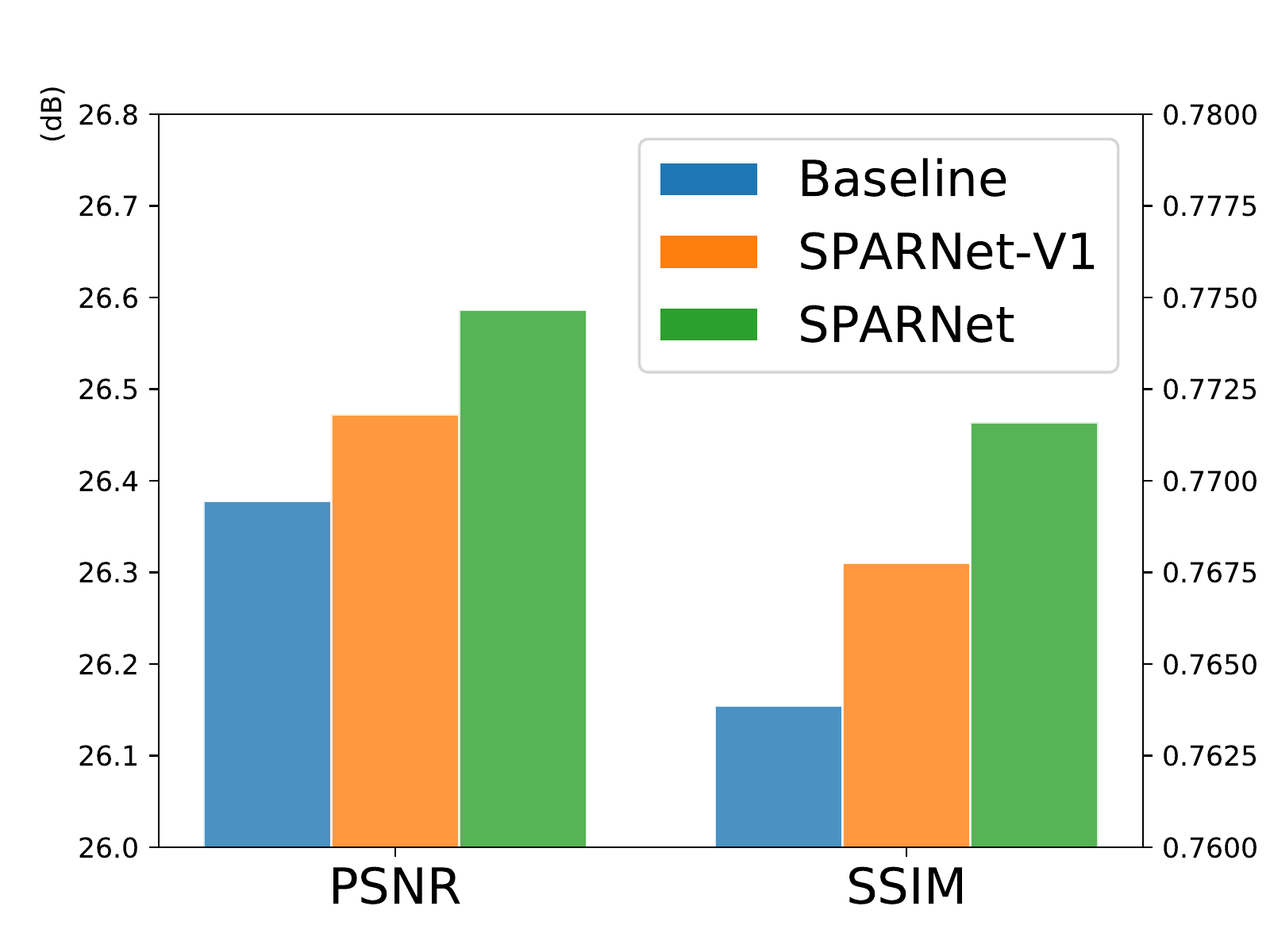}\label{fig:base-spar-quality}}
  \subfigure[Landmark Detection]{\includegraphics[width=.49\linewidth]{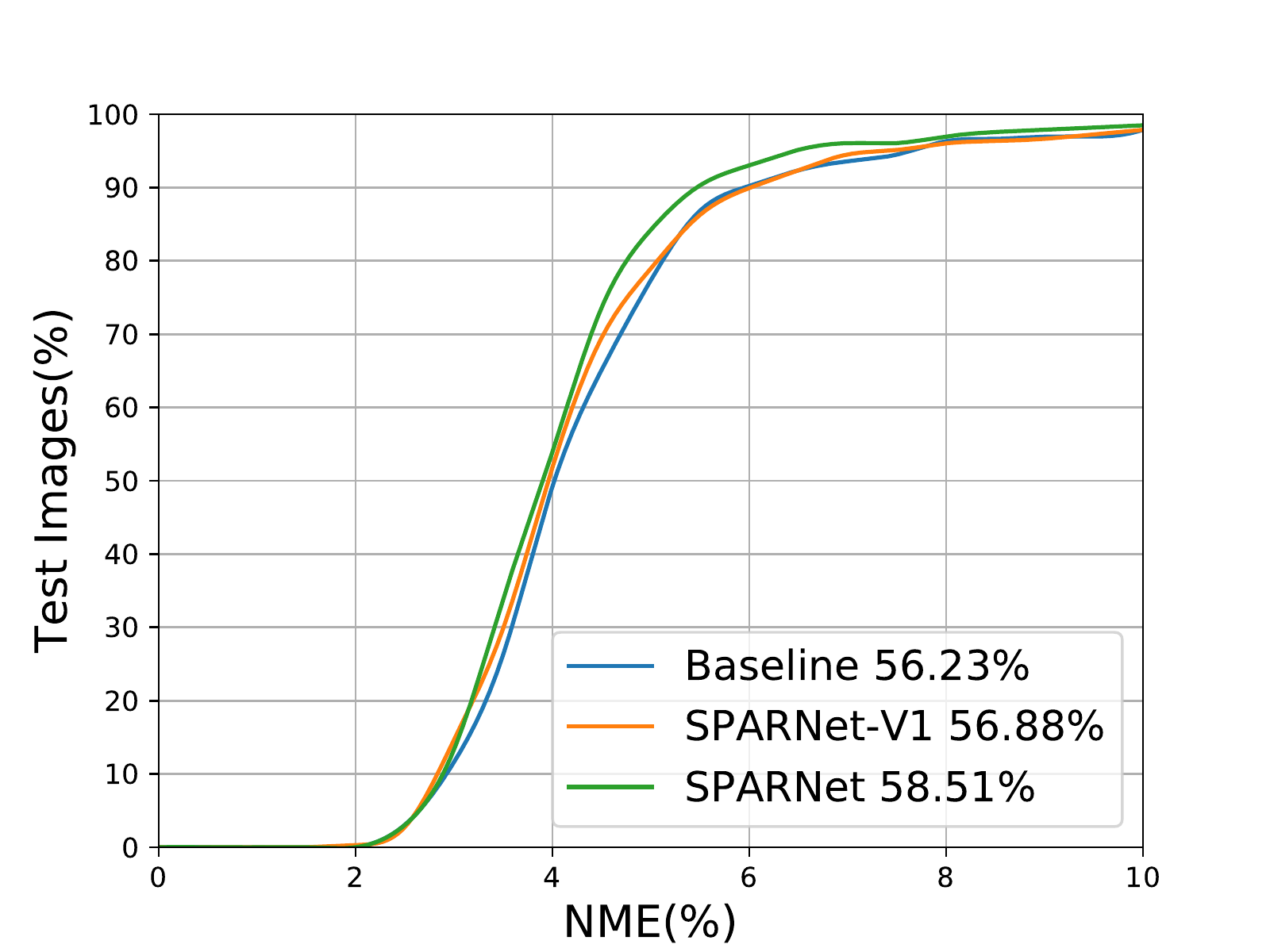}\label{fig:base-spar-lmk}}
  \subfigure[Visualization of the PSNR (left) and SSIM (right) error maps. The heatmaps for Baseline and SPARNet are the average 2D PSNR/SSIM error maps (larger value indicates larger error), and the improvement heatmaps show the error reduction (larger value indicates more error reduction).]{\includegraphics[width=.99\linewidth]{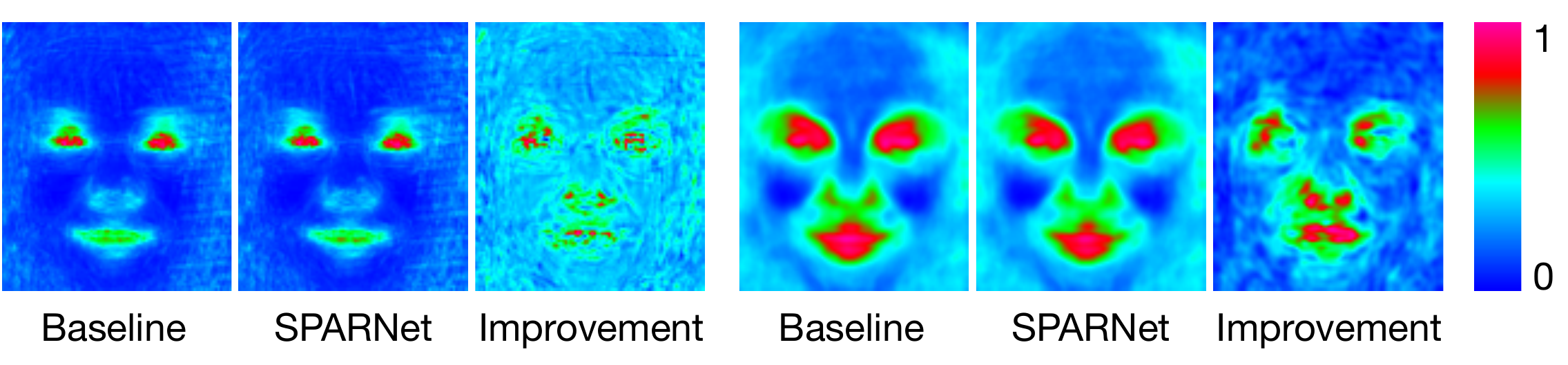} \label{fig:score-improve-visual}}
  \caption{Quantitative comparison between Baseline, SPARNet-V1 and SPARNet on the Helen test set.} \label{fig:base-spar}
\end{figure}

\subsubsection{Ablation Study}

We compare the following three variants of our model:

\begin{itemize}
  \item Baseline: residual SR network without any spatial attention branches.
  \item SPARNet-V$N$: To evaluate the effectiveness of using multiple FAU blocks, we keep the feature branch unchanged and vary the numbers of attention branch used in SPARNet \footnote{Reducing the numbers of entire FAU would make the network shallower which would definitely make the results worse, therefore we only vary the number of attention branch}. Considering that decoder parts are supposed to be more sensitive to spatial attention, we gradually increase the number of attention branches from the backend. We denote models with $N$ spatial attention branches as SPARNet-V$N$, where $N \in \{1, 2, 4, 8, 16\}$.  
  \item SPARNet-S$M$: To demonstrate that it is crucial for the attention mechanism to learn from multi-scale features, we vary the numbers of downsample/upsample blocks to change the smallest size (scale) of feature maps in the bottleneck. We denote SPARNet with $M\times M$ size of feature maps in the bottleneck of attention branch as SPARNet-S$M$, where $M \in \{2, 4, 8, 16\}$ and smaller $M$ indicates more scales of features are used.  
  \item SPARNet: the full model used in this work, \ie, SPARNet-V16-S4. 
\end{itemize}

\begin{figure}[t]
  \centering
  \includegraphics[width=.99\linewidth]{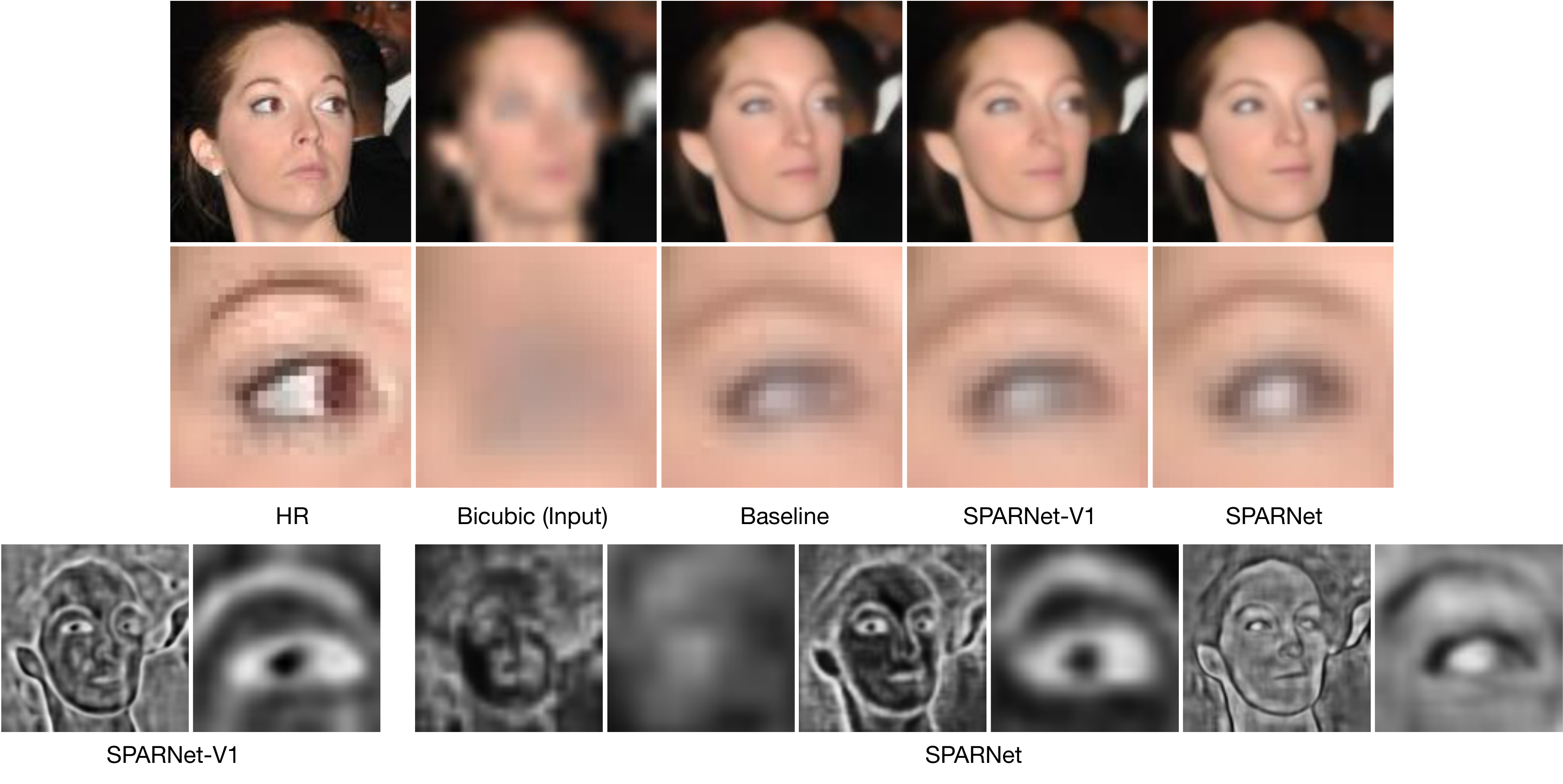} 
  \caption{Results on the Helen test set. Top: HR photo and SR results; bottom: spatial attention map in FAU of SPARNet-V1 (left) and last three FAUs of SPARNet (right).} \label{fig:ablation-result-example}
\end{figure}

\para{Effectiveness of Spatial Attention} To evaluate the effectiveness of the proposed spatial attention mechanism, we compare the results of SPARNet-V1 and SPARNet with that of the Baseline model. Fig~\ref{fig:base-spar-quality} shows that, with the proposed spatial attention mechanism, we can generate results with better image quality in terms of both PSNR and SSIM scores. This is expected as the spatial attention mechanism enables our network to focus on \blank{and} better recover the key \blank{face} structures. To visualize the improvement brought by the attention mechanism, we align the faces to a fixed template based on five key facial landmarks (\ie two eye centers, nose, and two mouth corners), calculate 2D PSNR and SSIM error maps\footnote{The PSNR error is calculated by mean square error, and the SSIM error is calculated by subtracting the SSIM score from 1.} for each image, and average the error maps over the whole test set. It can be observed from Fig.~\ref{fig:score-improve-visual} that pixels corresponding to key \blank{face} structures are most difficult to recover (\ie, with larger errors), and as expected, most of the improvement happens around them. Furthermore, we compare the landmark detection results on the super-resolved images. Better recovery of key \blank{face} structures should lead to \blank{higher} landmark detection accuracy. It can be seen in Fig.~\ref{fig:base-spar-lmk} that SPARNet achieves better landmark detection accuracy ($58.51\%$ AUC) than the Baseline ($56.23\%$ AUC). This demonstrates the benefit of the spatial attention mechanism. Fig.~\ref{fig:ablation-result-example} shows some SR results on the Helen test set. It can be observed that SPARNet can produce sharper and clearer \blank{face} structures than the Baseline, especially for the eyes.

\begin{figure}[ht]
    \centering
    \centering
    \subfigure[Evaluation PSNR score of SPARNet-V$N$ during training.]{\includegraphics[width=.49\linewidth]{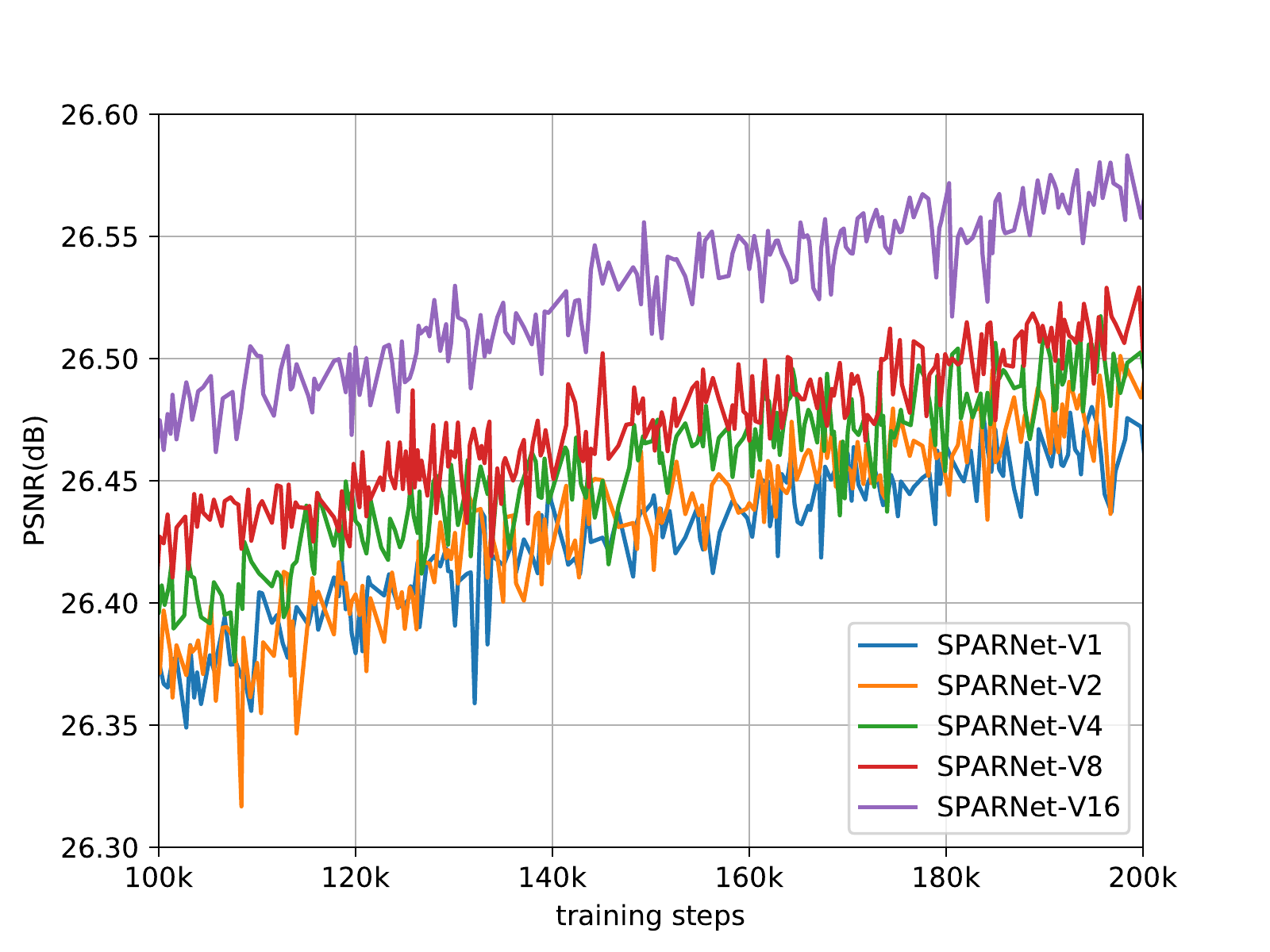}\label{fig:ablation-faus}}
    \subfigure[Evaluation PSNR score of SPARNet-S$M$ during training.]{\includegraphics[width=.49\linewidth]{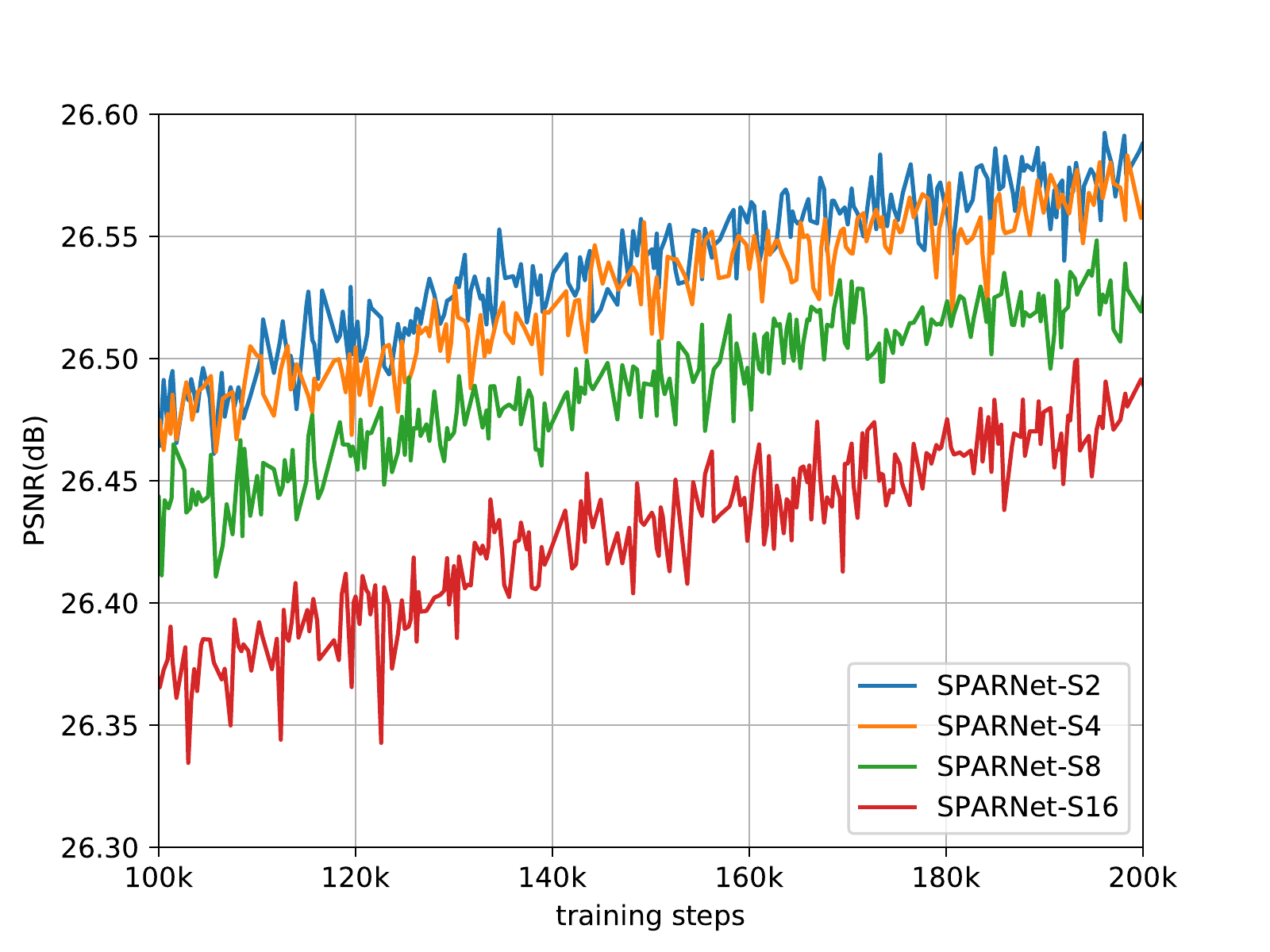}\label{fig:ablation-hourglass}}
    \caption{Ablation study of SPARNet-V$N$ and SPARNet-S$M$ on Helen testset. We set $M=4$ when training SPARNet-V$N$, and $N=16$ for SPARNet-S$M$.}
    \label{fig:more-sparnet-ablatoin}
\end{figure}

\para{Number of FAUs} By comparing the results of SPARNet-V1 with that of SPARNet in Fig.~\ref{fig:base-spar}, we can see that SPARNet, which is composed entirely of FAUs, performs better than SPARNet-V1, which has only one FAU at the end.  Although the attention map of SPARNet-V1 can focus on the key \blank{face} structures, some noises also show up on the face region \blank{and cause} some distraction \blank{(see Fig.~\ref{fig:ablation-result-example} bottom left)}. This suggests that bootstrapping the key \blank{face} structures using a single FAU may be sub-optimal. On the other hand, the \blank{sequence of} FAUs in SPARNet allow the attention maps to gradually focus on different key \blank{face} structures at different stages. For instances, one can see the attention map of the second last FAU focuses cleanly on the outline of the key \blank{face} structures, and that of the last FAU focuses on the general face region \blank{(see Fig.~\ref{fig:ablation-result-example} bottom right). In Fig. \ref{fig:ablation-faus}, we show more results comparison of SPARNet-V$N$. It can be observed that the evaluation PSNR scores of SPARNet-V$N$ show a positive correlation with $N$, which demonstrates the effectiveness of stacking multiple FAUs.}  

\para{Multiscale features in FAU} The results of SPARNet-S$M$ are shown in Fig. \ref{fig:ablation-hourglass}. We can see that models with more scales (\ie, smaller $M$) generally perform better, but the improvement declines when $M$ becomes smaller. The performance of SPARNet-S$4$ is similar to SPARNet-S$2$. Therefore, we set $M=4$ for SPARNet to balance the performance and computation cost.

\subsection{Comparisons of SPARNet against Other Methods}

We compare SPARNet with state-of-the-art general image SR method RCAN~\cite{zhang2018rcan}, latest attention network for image classification CBAM~\cite{woo2018cbam}\blank{,} and other face SR methods, including URDGN~\cite{yu2016ultra}, Wavelet-SRNet~\cite{huang2017wavelet}, Attention-FH~\cite{Cao2017Attention}, FSRNet~\cite{Chen2018CVPR}\blank{,} SICNN \cite{zhang2018super}, PFSRNet \cite{pfsrnet} and DICNet \cite{dicnet}. When comparing with RCAN and CBAM, we replace our FAU with the corresponding attention module (borrowed from public released codes) and keep network depth the same for a fair comparison. \emph{All models are trained \blank{using} the same dataset except for FSRNet and SICNN.} Note that FSRNet only provides test code and SICNN relies on a much larger pre-aligned training set. 

\subsubsection{Overall Results}
\begin{figure*}[ht]
  \includegraphics[width=.99\linewidth]{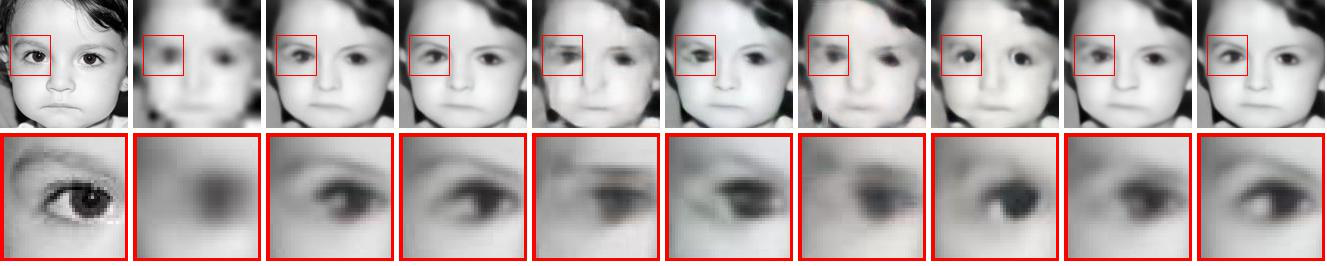}
  \includegraphics[width=.99\linewidth]{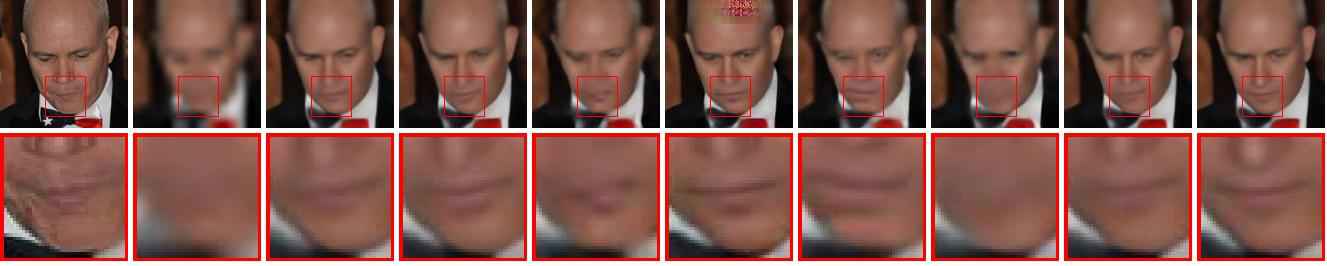}
  \begin{minipage}[t]{.99\textwidth}
    \footnotesize
    \centering
    \begin{tabular}{*{11}{C{1.4cm}}}
    (a) HR & (b) Bicubic & (c) RCAN~\cite{zhang2018rcan} & (d) CBAM~\cite{woo2018cbam} & (e) URDGN~\cite{yu2016ultra} & (f) Wavelet~\cite{huang2017wavelet} & (g) Att-FH~\cite{Cao2017Attention} & (h) FSRNet~\cite{yu2018face} & (i) Baseline & (j) SPARNet 
    \end{tabular}
  \end{minipage}
  \caption{Qualitative comparison with state-of-the-art methods. The resolution of the input is $16\times16$ and the upscale factor is $8\times$. }
  \label{fig:qual-comp}
\end{figure*}

Quantitative comparison with other state-of-the-art methods on the Helen test set is shown in Table~\ref{tab:psnr-ssim}. SPARNet clearly outperforms both generic SR methods and face SR methods in terms of both PSNR and SSIM scores. With the spatial attention mechanism, SPARNet can better recover important \blank{face} structures. One can observe from Fig.~\ref{fig:qual-comp} that while most of the other methods fail to recover the eyes and nose, and the shapes are blurry, SPARNet can generate sharper results with shapes close to the ground truth HR images.

\begin{table*}[ht]
  \caption{Quantitative comparison on the Helen (first 3 rows) and UMD (last row) test \blank{sets} with a $16\times16$ input size and an upscale factor of $8\times$. The AUCs are calculated under a threshold of $10\%$. The results of FSRNet$^*$ are generated using the test model provided by its authors.}
  \label{tab:psnr-ssim}
  \centering
  \begin{tabular}{|c|ccccccc|cc|}
    \hline
    Method & Bicubic & RCAN & CBAM & URDGN & Wavelet & Att-FH & FSRNet$^*$ & Baseline & SPARNet \\ \hline \hline
    PSNR  & 23.52  & 26.40  & \textcolor{blue}{\textbf{26.46}}  & 25.17  & 26.42  & 25.10  & 24.97 & 26.38 & \textcolor{red}{\textbf{26.59}}   \\ \hline
    SSIM  & 0.6408 & 0.7648 & 0.7666 & 0.7140 & \textcolor{blue}{\textbf{0.7711}} & 0.7188 & 0.7091 & 0.7639 & \textcolor{red}{\textbf{0.7716}}  \\ \hline
    AUC($<10\%$) & 4.42\% & 56.53\% & 56.83\% & 41.78\% & \textcolor{blue}{\textbf{58.44}}\% & 46.63\% & 40.42\% & 56.23\% & \textcolor{red}{\textbf{58.51\%}}  \\ \hline \hline
    Identity & 0.1851 & 0.5373 & \textcolor{blue}{\textbf{0.5378}} & 0.3981 & 0.5147 & 0.4731 & 0.4765 & 0.5302 & \textcolor{red}{\textbf{0.5546}}  \\ \hline
  \end{tabular}
\end{table*}

\subsubsection{Detailed Comparisons}

\begin{figure}[ht]
  \subfigure[SR results of CBAM and SPARNet.]{\includegraphics[width=.99\linewidth]{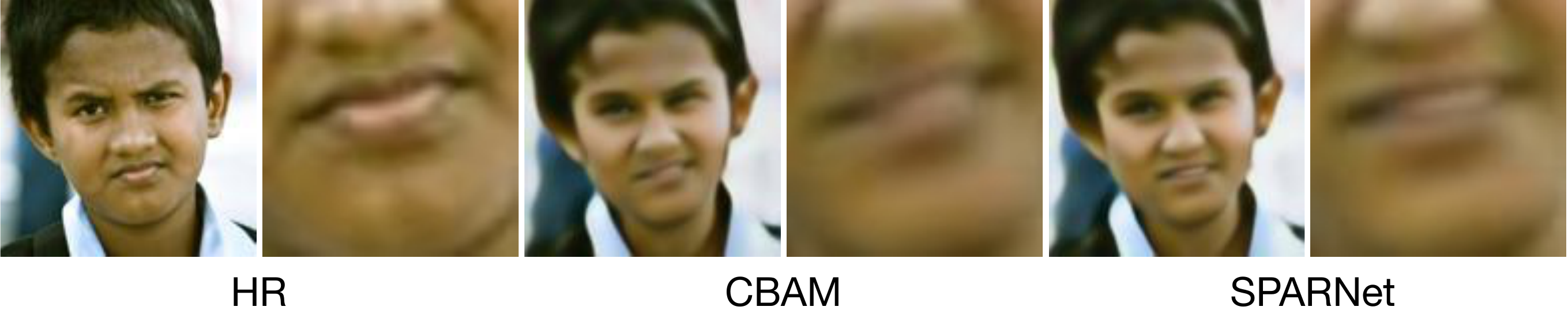} \label{fig:cbam-spar-a}}
  \subfigure[Attention maps of CBAM (top) and SPARNet (bottom).]{\includegraphics[width=.99\linewidth]{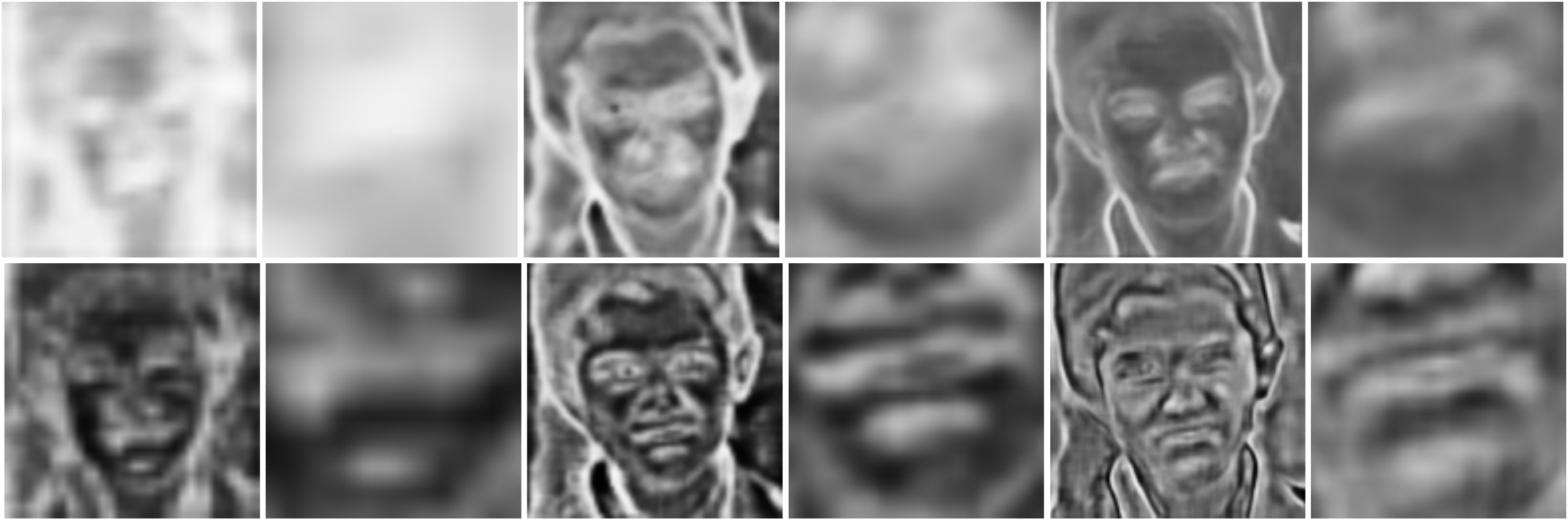} \label{fig:cbam-spar-b}}
  \caption{SR results and attention maps of CBAM and SPARNet.}
  \label{fig:cbam-spar}
\end{figure}

\para{Attention Mechanisms} Among all the methods under comparison, only RCAN, CBAM\blank{,} and Attention-FH have an attention mechanism. RCAN is designed for generic SR task. It applies a channel attention mechanism in the feature space. Although channel attention has been shown to be beneficial to general SR task, spatial structures are more important in very low resolution face super-resolution. This explains why RCAN does not show big improvement over the Baseline in terms of PSNR score ($26.40$ vs $26.38$). Attention-FH adopts a patch level attention and cannot recover details (see Fig~\ref{fig:qual-comp}(g)). CBAM also employs a spatial attention. However, it is manipulated to extract semantic information through pooling operations which cause loss of shape and edge details. We compare the SR \blank{results} and spatial attention maps generated by SPARNet and CBAM in Fig~\ref{fig:cbam-spar}. We can see that the attention maps produced by CBAM are blurry and not able to outline important details such as eyes and mouth, whereas the attention maps of SPARNet, thanks to the multi-scale feature based approach, provide more detailed structure information. Accordingly, the mouth and nose generated by SPARNet are sharper and clearer than that of CBAM. 

\para{Comparison with FSRNet~\cite{Chen2018CVPR}} Since FSRNet only provides test code, the results \blank{of FSRNet} reported in Table~\ref{tab:psnr-ssim} \blank{are} not obtained using the same training set as the others. To make a fair comparison, we retrain and test our model on the same dataset used by FSRNet (18K images from CelebA), and show the results in Table~\ref{tab:fsrnet-comp}. Although FSRNet uses face parsing map as an additional supervision, SPARNet still outperforms FSRNet in terms of PSNR and SSIM scores. This demonstrates the superiority of the proposed spatial attention mechanism.

\begin{table}[ht]
  \caption{Comparison between FSRNet and SPARNet on CelebA.} \label{tab:fsrnet-comp}
  \centering
  \begin{tabular}{|c|c|c|c|c|}
  \hline
  Method & FSRNet & SPARNet \\ \hline
  PSNR/SSIM & 26.31/0.7522 &  \textcolor{red}{\textbf{26.68/0.7741}} \\ \hline 
  \end{tabular}
\end{table}

\begin{figure}[t]
  \includegraphics[width=.99\linewidth]{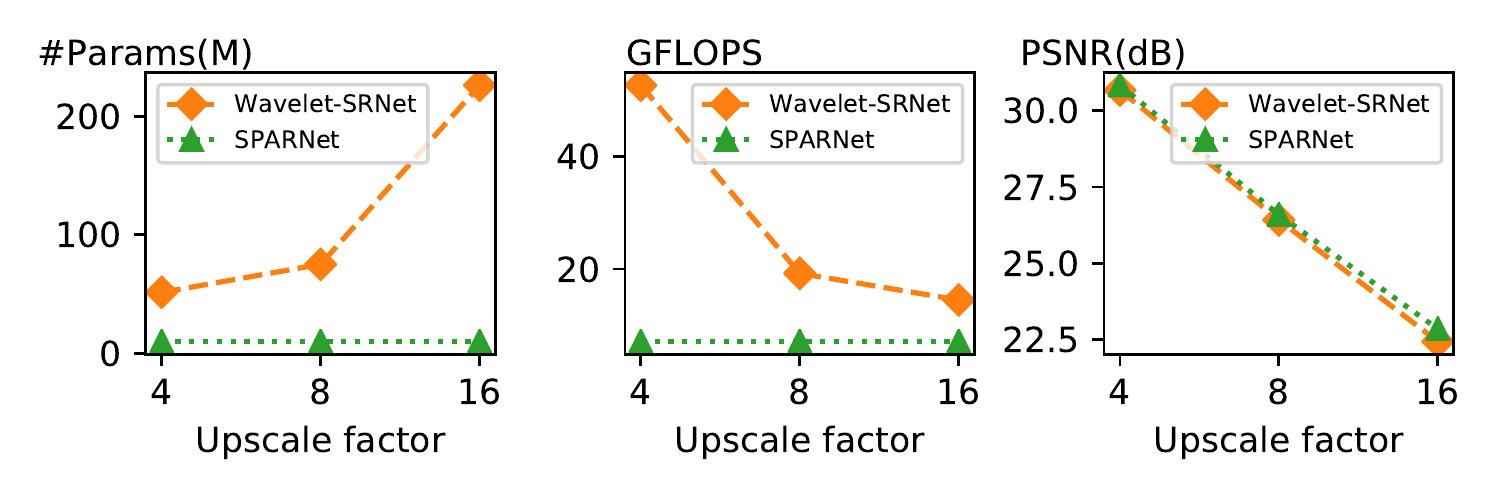}
  \caption{Performance and efficiency comparison between Wavelet-SRNet and SPARNet under different upscale factors.}
  \label{fig:wave-spar-comp}
\end{figure}

\para{Comparison with Wavelet-SRNet~\cite{huang2017wavelet}} Wavelet-SRNet shows closest performance to SPARNet in terms of SSIM score. Nonetheless, SPARNet requires much less parameters and is more computational efficient and flexible than Wavelet-SRNet. The performance and efficiency comparison between Wavelet-SRNet and SPARNet is summarized in Fig.~\ref{fig:wave-spar-comp}. Wavelet-SRNet has $4^n$ parallel subnets with an upscale factor of $2^n$. This implies the network parameters increase quadratically with the upscale factor. Besides, it requires different input size for each upscale factor. Because the output size is usually fixed, a smaller upscale factor needs a larger size of input when training the network. Since a larger input means more convolution operations and requires higher computational power, the GFLOPs of Wavelet-SRNet increase dramatically when the upscale factor is small. In contrast, SPARNet uses the same architecture for different upscale factors, and achieves better performance without increasing its computational complexity. 

\begin{figure}[t]
    \centering
    \includegraphics[width=0.99\linewidth]{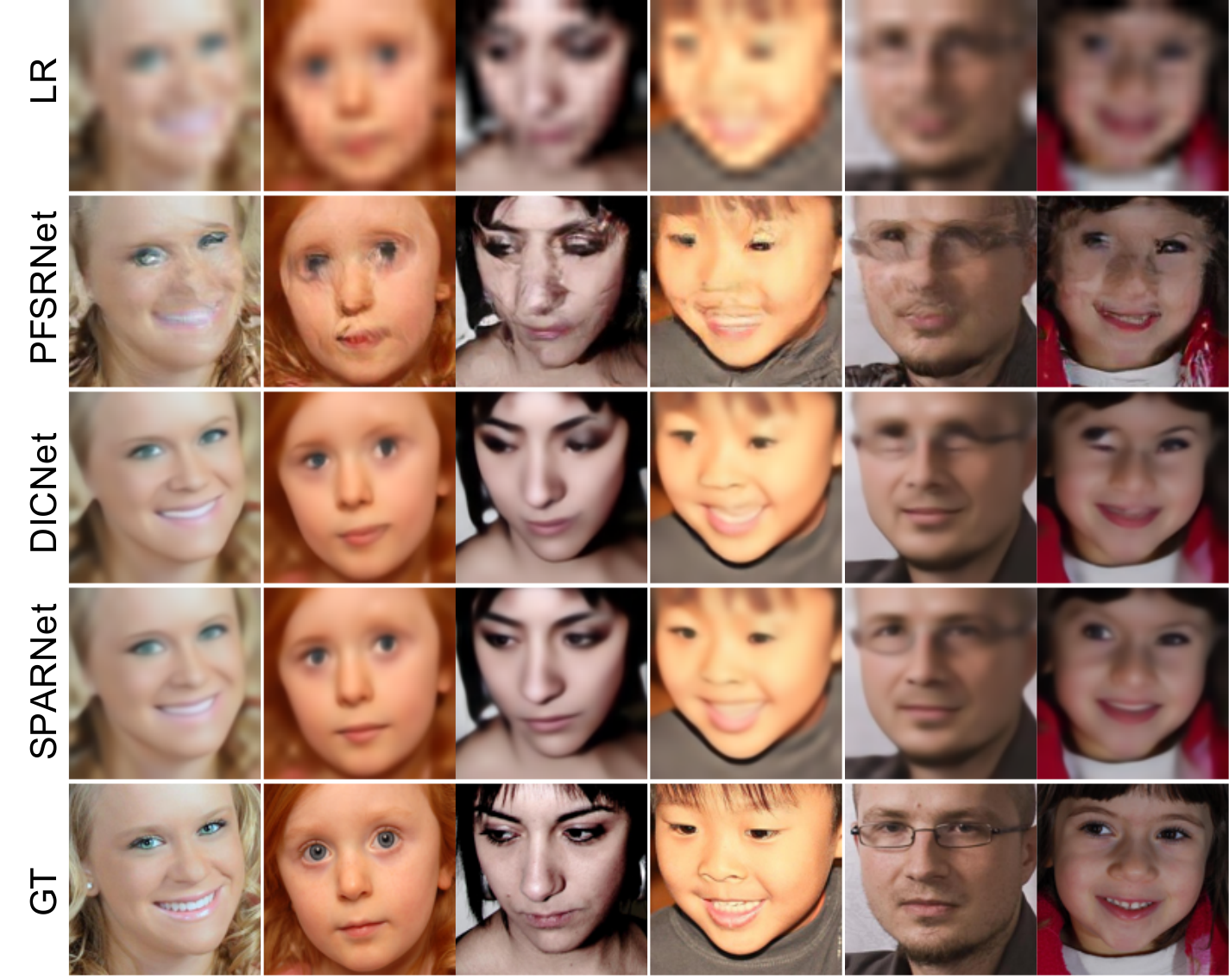}
    \caption{Visual comparison between PFSRNet, DICNet, and SPARNet on Helen test dataset provided by DICNet. All results are generated by public models trained on CelebA. PFSRNet generates bad results for unaligned test faces. SPARNet generates better key face components than DICNet especially in the eyes.}
    \label{fig:extra-comp}
\end{figure}

\para{Comparison with landmark based methods} We compare SPARNet with two recent landmark based methods PFSRNet\footnote{\url{https://github.com/DeokyunKim/Progressive-Face-Super-Resolution}} \cite{pfsrnet} and DICNet\footnote{\url{https://github.com/Maclory/Deep-Iterative-Collaboration}} \cite{dicnet}. To get the best results of the compared method, we directly use their public models instead of retraining them since both of them are trained on CelebA, the same as SPARNet. We evaluate the performance on the Helen test dataset provided by DICNet because our test images are cropped in a different way and DICNet performs bad on it. The quantitative results and visual examples are shown in Table \ref{tab:extra-comp} and Fig. \ref{fig:extra-comp} respectively. As we can observe from Table \ref{tab:extra-comp}, SPARNet shows the best PSNR and SSIM scores in the test dataset. The visual results in Fig. \ref{fig:extra-comp} indicate that SPARNet can recover the key face components, especially the eyes, better than DICNet even without any extra landmark information. We hypothesize that this is because it is too difficult to detect accurate landmarks for low resolution face images, and the multi-stage iterative process in DICNet amplifies the error. As for PFSRNet, we found it performs badly when test faces are not aligned. Since detecting landmarks from $16\times16$ LR face is by itself a difficult task, we find the practical value of PFSRNet is limited compared with DICNet and our SPARNet. We also show the computation complexity in Table \ref{tab:extra-comp}. Compared with state-of-the-art method DICNet, our method produces better results with much less parameters. Moreover, our SPARNet is faster than DICNet which is an iterative framework. Compared with PFSRNet, our method achieves much better performance with similar computation complexity.

\begin{table}[t]
    \centering
    \caption{Quantitative comparison on the Helen test dataset provided by DICNet \cite{dicnet}.}
    \label{tab:extra-comp}
    \begin{tabular}{|c|c|c|c|}
    \hline
      Methods & PFSRNet \cite{pfsrnet} & DICNet \cite{dicnet} & SPARNet \\ \hline
      PSNR & 24.13 & 26.73 & \redb{26.97} \\ \hline
      SSIM & 0.6688 & 0.7955 & \redb{0.8026} \\ \hline \hline
      Params (M) & 8.97 & 22.80 & 9.86 \\ \hline
      Time (s) & 0.045 & 0.065 & 0.051 \\ \hline
    \end{tabular}
\end{table}

\para{Comparison with Different Upscale Factors} We also trained models with $4\times$, and $16\times$ upscale factors. The $4\times$ model was trained from scratch, whereas the $16\times$ model was initialized with the pre-trained $8\times$ model. We show the results of $4\times$ and $16\times$ upscale factors in Table~\ref{tab:multi-res}. It can be seen that SPARNet achieves state-of-the-art results for both $4\times$ and $16\times$ upscale factors, especially for \blank{the} $16\times$ upscale factor. 

\begin{table}[!thb]
  \caption{Image quality comparison on Helen with upscale factors $4\times$ and $16\times$}
  \label{tab:multi-res}
  \setlength{\tabcolsep}{0.25em} 
  \centering
  \begin{tabular}{|c|c|cccc|c|}
  \hline
    & Scale & Bicubic & RCAN & CBAM & Wavelet & SPARNet \\ \hline \hline 
  PSNR &  \multirow{2}{*}{$4\times$} & 27.43 & 30.73 & \textbf{\textcolor{blue}{30.76}} & 30.67 & \textbf{\textcolor{red}{30.83}} \\ 
  SSIM &  & 0.8049 & 0.8857 & 0.8861 & \textbf{\textcolor{red}{0.8888}} & \textbf{\textcolor{blue}{0.8872}} \\ \hline \hline
  PSNR &  \multirow{2}{*}{$16\times$} & 20.22 & 20.71 & \textbf{\textcolor{blue}{20.75}} & 22.44 & \textbf{\textcolor{red}{22.85}}\\ 
  SSIM &  & 0.5180 & 0.6360 & \textbf{\textcolor{blue}{0.6375}} & 0.6314 & \textcolor{red}{\textbf{0.6426}} \\ \hline 
  \end{tabular}
\end{table}

\subsubsection{Landmark Detection Results}
As mentioned previously, the proposed spatial attention mechanism helps recover important \blank{face} structures. Landmark detection can be exploited to evaluate such an ability. From Table~\ref{tab:psnr-ssim}, we can see that SPARNet \blank{achieves} the best performance in landmark detection. Wavelet-SRNet's performance is similar to SPARNet, but SPARNet is more efficient (refer to Fig.~\ref{fig:wave-spar-comp}) and shows much better performance (in terms of PSNR/SSIM scores) for \blank{the} $16\times$ upscale factor (refer to Table~\ref{tab:multi-res}).  

\subsubsection{Identity Similarity Results}
We first evaluate the models in Table~\ref{tab:psnr-ssim} on the UMD-Face test set. Note that these models are trained on CelebA without pre-alignment. SPARNet achieves the best performance. We also notice that GAN based methods such as URDGN \blank{are} worse than non-GAN based methods. This is because GAN based methods \blank{target} at producing realistic textures but \blank{they do} not care whether the generated textures \blank{are} consistent with the \blank{HR images}. Therefore, \blank{they} may generate textures which look sharper but disturb the identity information. To further explore how well SPARNet can preserve identity information, we retrain our model using the same dataset as SICNN to allow a fair comparison. We denote this model as SPARNet$^*$. From Table~\ref{tab:id-sim}, we can see that SPARNet$^*$ shows better performance than SICNN even without using any explicit identity supervision. This proves that the proposed spatial attention mechanism is beneficial to preserving identity information in face SR.  

\begin{table}[thb]
  \caption{Comparison of identity similarity between SICNN and SPARNet$^*$. (SPARNet$^*$ is trained with the same dataset as SICNN.)}
  \label{tab:id-sim}
  \centering
  \begin{tabular}{|c|c|c|c|}
  \hline
  Method  & SICNN~\cite{zhang2018super} & SPARNet$^*$ \\ \hline \hline 
  Identity Similarity & 0.5978 & \textcolor{red}{\textbf{0.6272}} \\ \hline 
  \end{tabular}
\end{table}

\subsection{Evaluation of SPARNetHD}

Different from SPARNet which aims to hallucinate very low resolution faces (\ie, $16\times16$) that are difficult to recognize, the extended SPARNetHD tries to deal with real world low resolution faces. \blank{Typical} application scenarios \blank{include} old photos \blank{and} unclear faces shot by \blank{low-end} devices. These LR faces usually \blank{do} not have fixed upscale factors and are usually noisy and blurry. \blank{Hence}, we need \blank{a} different degradation model \blank{for data synthesis. We} also need high resolution datasets in order to get high resolution outputs.     

\subsubsection{Degradation Model}
According to previous works \cite{Li_2018_ECCV,xu2017learning} and common practice in \blank{single image SR} framework, we generate the LR image $I_l^s$ \blank{from the HR image $I_h$ using} the following degradation model:
\begin{equation}
  I_l^s = ((I_h \blank{*} \textbf{k}_\varrho)\downarrow_s + \textbf{n}_\delta)_{JPEG_q}, \label{equ:degrade} 
\end{equation}
where \blank{$*$} represents the convolution operation between \blank{$I_h$} and a blur kernel $\textbf{k}_\varrho$ with parameter $\varrho$. $\downarrow_s$ is the downsampling operation with a scale factor $s$. $\textbf{n}_\delta$ denotes the additive white Gaussian noise (AWGN) with a noise level $\delta$. $(\cdot)_{JPEG_q}$ indicates the JPEG compression operation with quality factor $q$. The hyper parameters $\varrho, s, \delta, q$ are randomly selected for each HR image $I_h$, and $I_l^s$ is generated online. We set the hyper parameters as 
\begin{itemize}
\item $\textbf{k}_\varrho$ is the blur kernel. We randomly choose one of the following four kernels: Gaussian Blur \blank{($3 \leq \varrho \leq 15$)},  Average Blur \blank{($3 \leq \varrho \leq 15$)}, Median Blur \blank{($3 \leq \varrho \leq 15$)}, Motion Blur \blank{($5 \leq \varrho \leq 25$)};
\item $\downarrow_s$ is the downsample operation. The scale factor $s$ is randomly selected in $[\frac{16}{512}, \frac{128}{512}]$; 
\item $\textbf{n}_\delta$ is addictive white gaussian noise (AWGN) with \blank{$0 \leq \delta \leq 0.1\times255$};
\item $JPEG_q$ is the JPEG operation. The compression level is randomly chosen from $[60, 85]$, in which higher means stronger compression level.  
\end{itemize}
After \blank{obtaining} $I_l^s$, $I_{LR-UP}=(I_l^s)\uparrow_s$ \blank{serves as the LR input of SPARNetHD}. 

\subsubsection{Datasets and Evaluation Metrics}

\para{Training data} We adopt the FFHQ \cite{karras2019style} dataset \blank{for training}. This dataset consists of $70,000$ high-quality images at a size of $1024\times1024$ crawled from the internet. All images are automatically cropped and aligned. We resize the images to $512\times512$ with bilinear downsampling as the ground-truth HR images, and synthesize the LR inputs online with Eq. \ref{equ:degrade}. 

\para{Synthetic Test Data} We use the test partition of CelebAHQ dataset~\cite{karras2017progressive} as synthetic test data. CelebAHQ contains $30,000$ HR faces in total which are split into a training set ($24,183$ images), a validation set ($2,993$ images), and a test set ($2,824$ images). We generate the LR dataset with Eq.~\ref{equ:degrade}, denoted as CelebAHQ-Test.   

\para{Natural Test Data} We collect $1,020$ faces smaller than $48\times48$ from CelebA. We also collect some old photos from the internet for testing. All images are cropped and aligned in the same manner as FFHQ, and then resized to $512\times512$ using bicubic upsampling. We merge all these images together and create a new dataset containing $1,051$ natural LR faces, denoted as CelebA-TestN. 

\para{Evaluation Metrics} For CelebAHQ-Test, we use LPIPS (Learned Perceptual Image Patch Similarity) \cite{zhang2018perceptual} as our evaluation metric \blank{because it} can better represent the visual quality \blank{of} HR images than PSNR and SSIM. For CelebA-TestN, since there are no ground truth HR images, we use FID (Fréchet Inception Distance) \cite{heusel2017gans} to measure the similarity between the SR results and reference datasets containing HR face images. We use the HR version of CelebAHQ-Test as the reference dataset.  

\subsubsection{Ablation Study of Loss Functions}
\begin{figure*}[!ht]
  \centering
  \includegraphics[width=.99\textwidth]{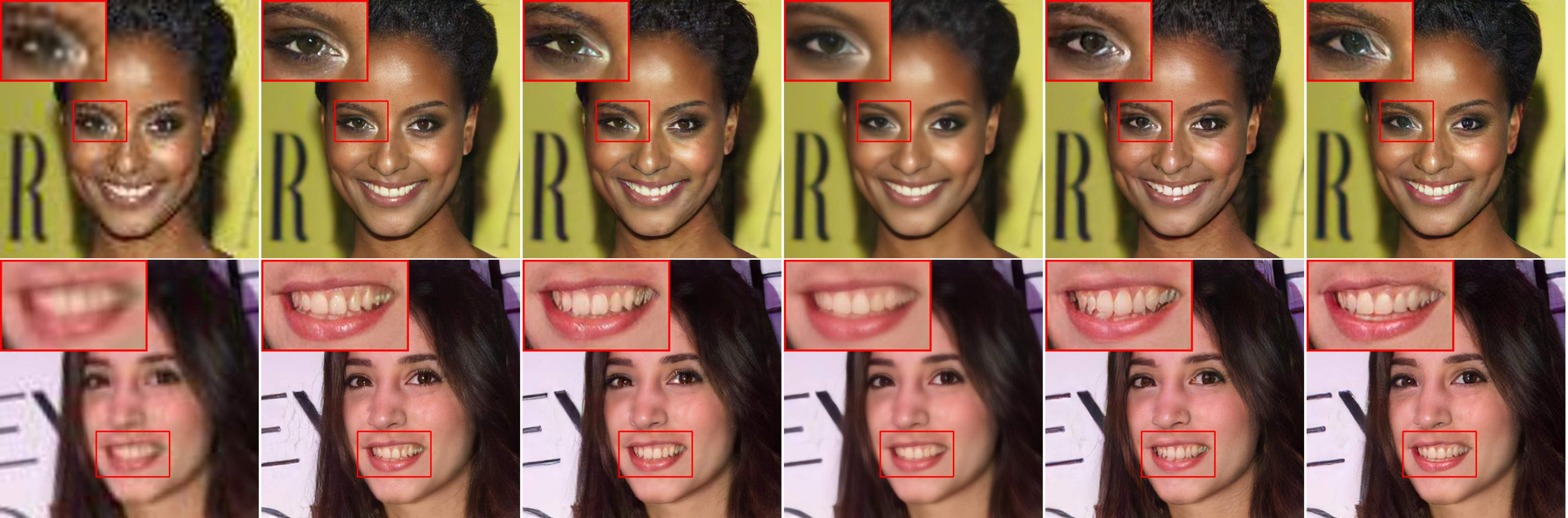}
  \begin{tabular}{*{6}{C{2.6cm}}}
    LR Input & SPARNetHD & w/o $\mathcal{L}_{pix}$ & w/o $\mathcal{L}_{GAN}$ & w/o $\mathcal{L}_{fm}$ & w/o $\mathcal{L}_{pcp}$ 
  \end{tabular}
  \caption{Ablation study of losses used in SPARNetHD on the synthetic dataset CelebAHQ-Test (first row) and real dataseet CelebA-TestN (second row). Better zoom in to see the details.}
  \label{fig:sparhd-ablation}
\end{figure*}
To verify the effectiveness of different loss terms in Eq. \ref{eq:loss-hd-final}, we conduct an ablation study by removing each of them separately. Same as SPARNetHD, all models are trained for 100k iterations which take about 3 days on a single GPU. The comparison results are shown in Table \ref{tab:quant-results} and Fig. \ref{fig:sparhd-ablation}. We can observe that $\mathcal{L}_{pix}$ has the least influence on the final performance because it mainly affects the subtle low level details. Comparing the results of second column and third column, the skin color of SPARNetHD (w/o $\mathcal{L}_{pix}$) is mixed with slight background green color compared with full SPARNetHD in the first row, and teeth shape is less natural w/o $\mathcal{L}_{pix}$ in the second row (please zoom in to see the details). From the 4th column, we can see that the network fails to generate sharp edges and realistic details without adversarial loss, and these greatly degrade the quantitative performance. In the 5th column, we can observe that the generated details are less realistic and there are some noise-like artifacts near the hair in the first row. This is mainly because the GAN network trained without $\mathcal{L}_{fm}$ is less stable. In the last column, we can see obvious shape distortions in the eyes and lips since $\mathcal{L}_{pcp}$ mainly helps to constrain mid-level and high-level semantics.

\subsubsection{Comparison with Other Methods}

\begin{figure*}[!ht]
  \centering
  \includegraphics[width=.99\textwidth]{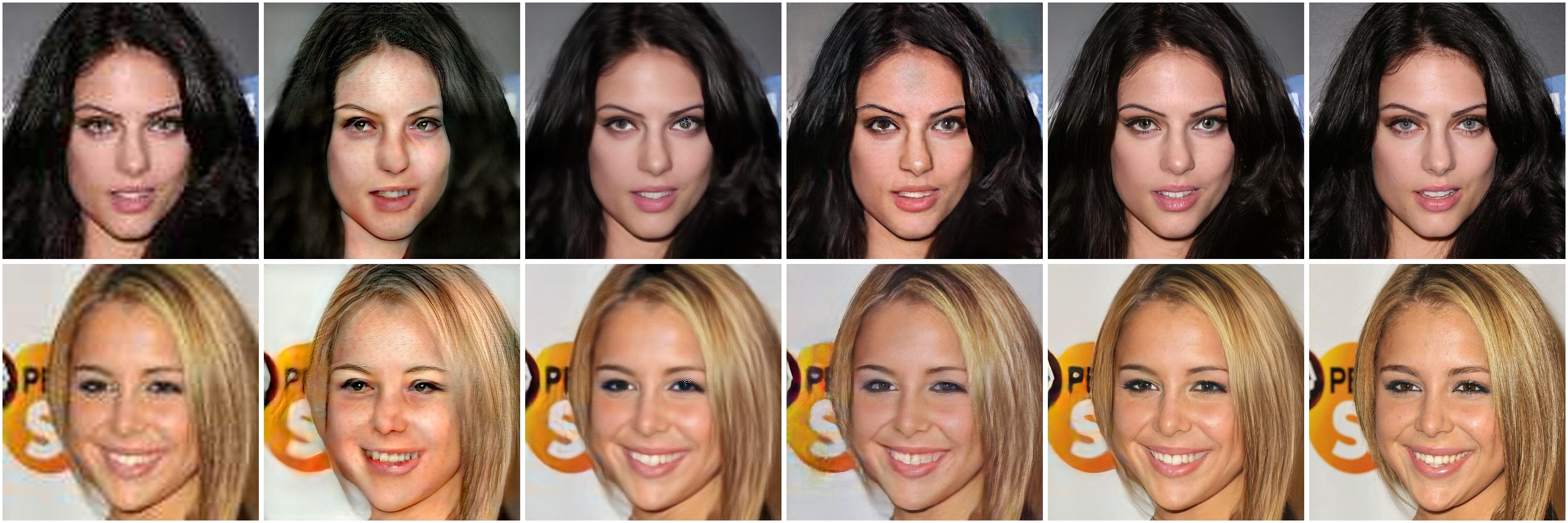}
  \begin{minipage}[t]{.99\textwidth}
    \centering
    \begin{tabular}{*{6}{C{2.6cm}}}
    (a) Synthetic LR & (b) ESRGAN & (c) Pix2PixHD & (d) SFTGAN & (e) Ours & (f) GT 
    \end{tabular}
  \end{minipage}
  \caption{Visual comparison between \blank{results produced by} SPARNetHD and other methods on \blank{the} synthetic dataset CelebAHQ-Test. Better zoom in to see the details.}
  \label{fig:sparhd-comp-synthetic}
\end{figure*}

\begin{figure*}[!ht]
  \centering
  \includegraphics[width=0.99\textwidth]{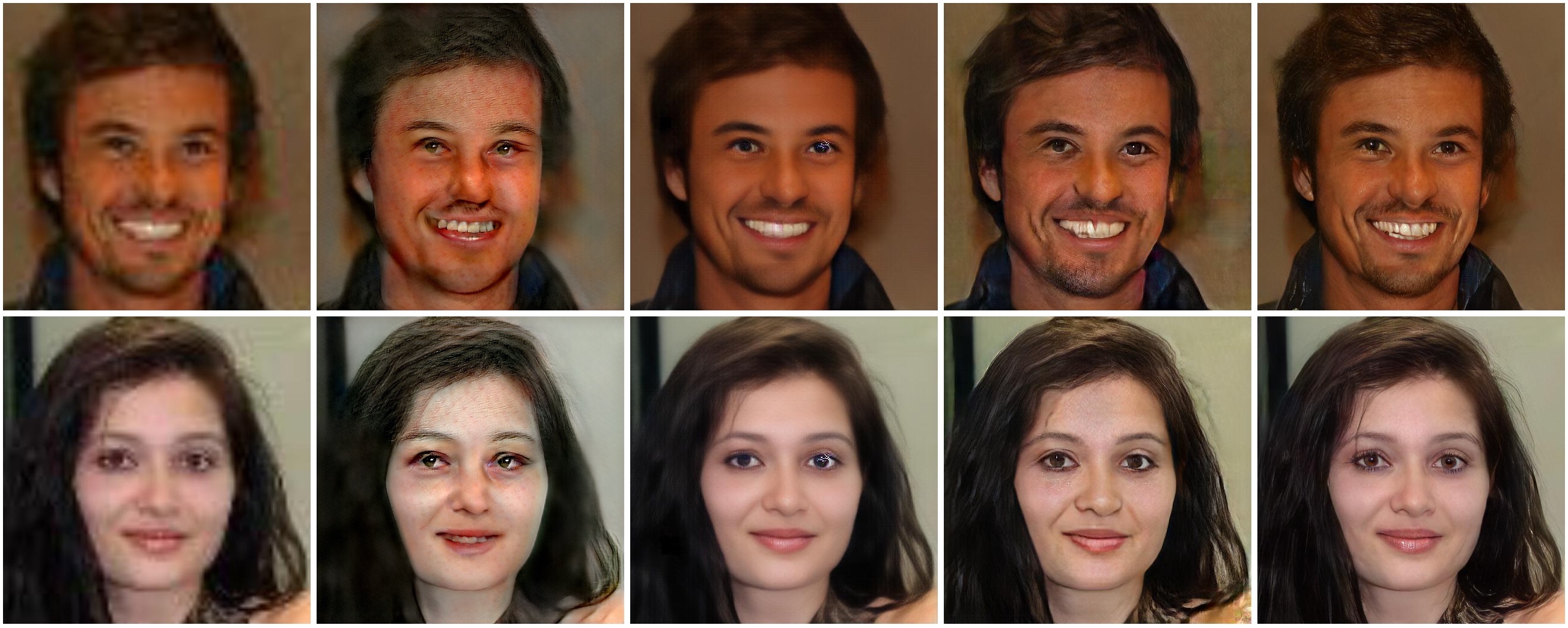}
  \begin{minipage}[t]{.99\textwidth}
    \centering
    \begin{tabular}{*{5}{C{3.1cm}}}
    (a) Real LR & (b) ESRGAN & (c) Pix2PixHD & (d) SFTGAN & (e) Ours  
    \end{tabular}
  \end{minipage}
  \caption{Visual comparison between \blank{results produced by} SPARNetHD and other methods on real LR dataset CelebA-TestN. Better zoom in to see the details.}
  \label{fig:sparhd-comp-natural}
\end{figure*}

We conduct experiments to compare SPARNetHD with other methods on CelebAHQ-Test and CelebA-TestN. \blank{Since very} few previous face SR methods can produce \blank{high resolution} outputs, we mainly compare SPARNetHD with general SR methods and GAN based image-to-image translation methods which \blank{provide} public codes and can be used to generate \blank{high resolution} outputs. Specifically, we compare SPARNetHD with state-of-the-art SR method ESRGAN~\cite{wang2018esrgan}, parsing map based SR method SFTGAN~\cite{wang2018sftgan}, high resolution image translation method Pix2PixHD \cite{wang2018pix2pixHD}, and blind face restoration method based on a reference image GFRNet \cite{Li_2018_ECCV}. We retrain ESRGAN, SFTGAN and Pix2PixHD. We use existing face parsing model to generate face parsing map for SFTGAN. As for GFRNet, we only \blank{carry out a} visual comparison with the results reported in the \blank{original} paper because it needs reference images for training and testing.   

\begin{table}[t]
\caption{Quantitative comparison on CelebAHQ-Test and CelebA-TestN. \emph{Note for w/o $\mathcal{L}_{GAN}$$^*$: because $L_{fm}$ depends on the discriminator, we also remove it when $\mathcal{L}_{GAN}$ is not used.}}
\label{tab:quant-results}
\centering
\begin{tabular}{l|c|c} 
\hline
Methods & CelebAHQ-Test (LPIPS$\downarrow$) & CelebA-TestN (FID$\downarrow$) \\
\hline\hline
ESRGAN \cite{wang2018esrgan} &  0.49 & 60.67 \\
SFTGAN \cite{wang2018sftgan}   &  0.36 & 37.76 \\
Pix2PixHD \cite{wang2018pix2pixHD} & 0.36 & 43.1  \\ 
\hline \hline
SPARNetHD & \redb{0.29} & \redb{27.16} \\
 w/o $\mathcal{L}_{pix} $ & 0.31 & 30.63 \\
 w/o $\mathcal{L}_{GAN}$$^*$ & 0.38 & 39.75 \\
 w/o $\mathcal{L}_{fm} $ & 0.32 & 33.52 \\
 w/o $\mathcal{L}_{pcp} $  & 0.31 & 32.97 \\
\hline
\end{tabular}
\end{table}

Table~\ref{tab:quant-results} gives \blank{a} quantitative \blank{comparison} on CelebAHQ-Test and CelebA-TestN. We can observe that SPARNetHD \blank{outperforms the other methods on both datasets}. Pix2PixHD performs better than ESRGAN because of the multi-scale discriminators. \blank{The} fact that the results of SPARNetHD are much better than \blank{that of} Pix2PixHD demonstrates the effectiveness of the proposed attention mechanism\blank{,} which is the key difference between SPARNetHD and Pix2PixHD. SFTGAN shows slightly better performance than Pix2PixHD with the help of face parsing maps. This \blank{shows the advantage of taking parsing map as additional supervision}. Nevertheless, the proposed SPARNetHD outperforms SFTGAN by a large margin. This is because parsing map can only provide coarse semantic level guidance, while the proposed spatial attention mechanism cannot only provide coarse semantic guidance but also give low level texture guidance.    

The \blank{qualitative} results in Fig.~\ref{fig:sparhd-comp-synthetic} and Fig~\ref{fig:sparhd-comp-natural} \blank{are consistent} with the observations in Table~\ref{tab:quant-results}. Fig.~\ref{fig:sparhd-comp-synthetic} shows the results on \blank{the} synthetic test dataset CelebAHQ-Test. We can see that ESRGAN fails to generate realistic faces. While the results of Pix2PixHD \blank{look} much better, \blank{undesirable artifacts show up} in the right eyes. SFTGAN does not have such problem\blank{, but it cannot} generate detailed textures\blank{, especially in the hair and teeth}. In contrast, SPARNetHD can restore key \blank{face} components as well as the texture details in the hair and teeth. \blank{Similar} phenomenon can also be observed \blank{in} Fig.~\ref{fig:sparhd-comp-natural}\blank{,} which \blank{shows} the results on natural LR faces. While the competitive methods generate many artifacts for real LR faces, our results are much more robust and \blank{natural. This} illustrates the good generalization ability of SPARNetHD. 

\begin{figure}[!htb]
  \centering
  \includegraphics[width=.99\linewidth]{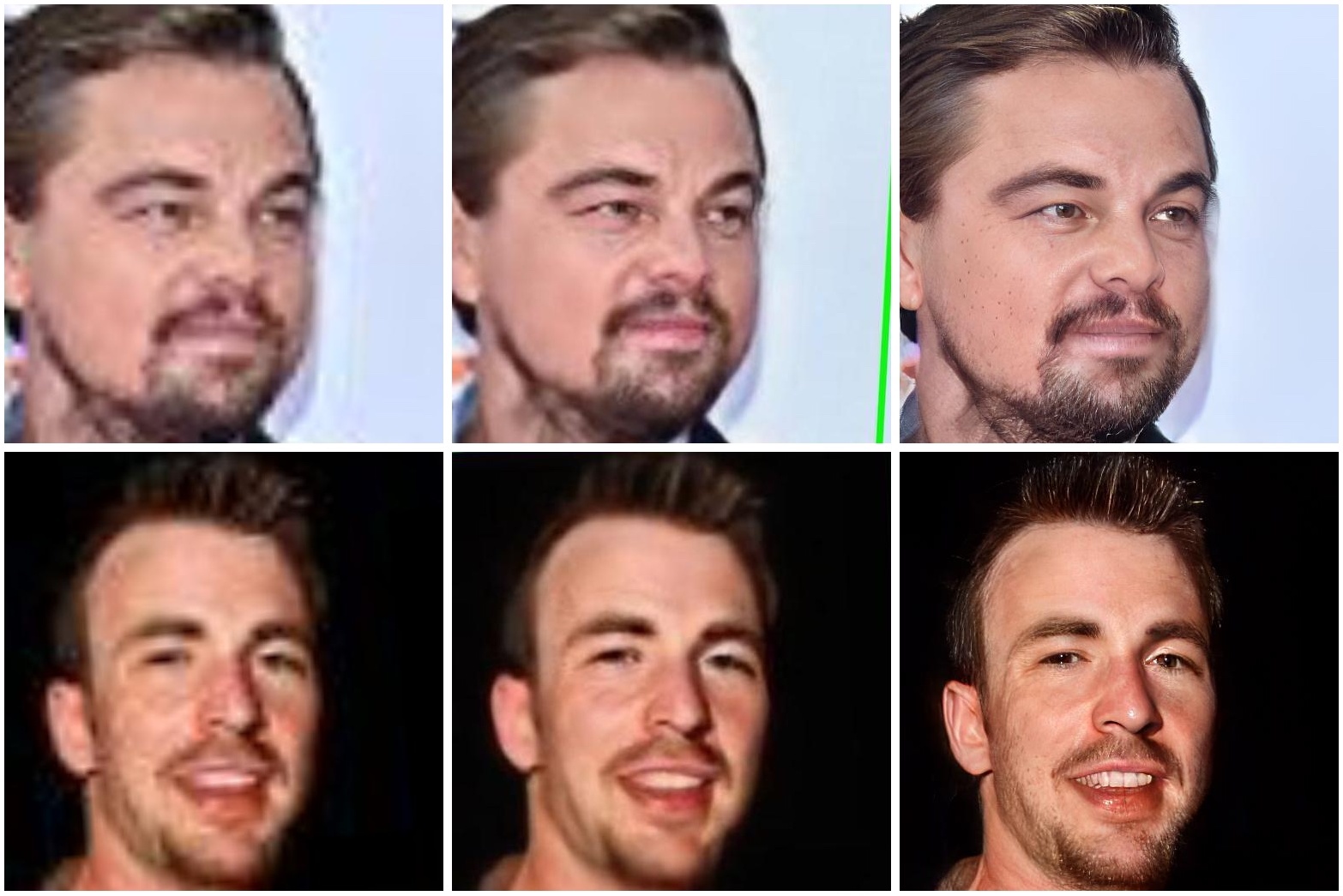}
  \begin{minipage}[t]{.99\linewidth}
    \centering
    \begin{tabular}{*{3}{C{2.6cm}}}
    (a) Real LR & (b) GFRNet & (c) Ours
    \end{tabular}
  \end{minipage}
  \caption{Visual comparison between \blank{results produced by} SPARNetHD and GFRNet. Better zoom in to see the details.}
  \label{fig:sparhd-comp-gfrnet}
\end{figure}

We also compare SPARNetHD with a recent blind face restoration method GFRNet. GFRNet takes a LR image and a HR image of the same person as \blank{input} to restore the LR image. We \blank{show some qualitative results} on natural LR faces in Fig.~\ref{fig:sparhd-comp-gfrnet}. It can be observed that SPARNetHD generates much better texture details than GFRNet even without \blank{utilizing} any additional information \blank{in} the training stage. 

\section{Conclusion}
We propose a SPatial Attention Residual Networks (SPARNet) for very low resolution face super-resolution. SPARNet is composed \blank{by stacking} Face Attention Units (FAUs), which extend vanilla residual block with a spatial attention branch. The spatial attention mechanism allows the network to \blank{pay less attention on the less feature-rich regions.} 
\blank{This} makes the training of SPARNet more effective and efficient. Extensive experiments with various kinds of metrics illustrate the advantages of SPARNet over current state-of-the-arts. We further extend SPARNet to SPARNetHD with more channel numbers and multi-scale discriminator networks. SPARNetHD trained on synthetic datasets is able to generate realistic and \blank{high resolution} outputs (\ie, $512\times512$) for LR face images. Quantitative and qualitative comparisons with other methods indicate that the proposed spatial attention mechanism is beneficial to restore texture details of LR face images. We also demonstrate \blank{that} SPARNetHD can \blank{generalize} well to real world LR faces, \blank{making it highly practical and applicable}.  




\section*{Acknowledgments}

The authors would like to thank Tencent AI Lab for supporting this research. The work of Kwan-Yee K. Wong was supported by a grant from the Research Grant Council of the Hong Kong (SAR), China, under the Project HKU 17203119.

\ifCLASSOPTIONcaptionsoff
  \newpage
\fi

\bibliography{bibtex/bib/egbib.bib}
\bibliographystyle{IEEEtran}

%

\begin{IEEEbiography}[{\includegraphics[width=1in,height=1.25in,clip,keepaspectratio]{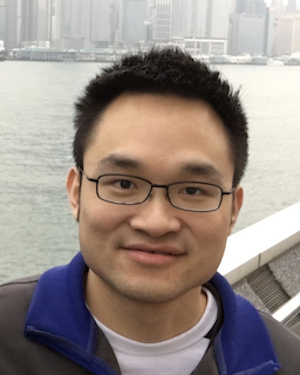}}]{Chaofeng Chen}
is currently a Ph.D. student in the Department of Computer Science, The University of Hong Kong. He received his B.E. degree from Huazhong University of Science and Technology in 2015.  His research interests are centered around computer vision and deep learning. 
\end{IEEEbiography}
\vspace{-1em}

\begin{IEEEbiography}[{\includegraphics[width=1in,height=1.25in,clip,keepaspectratio]{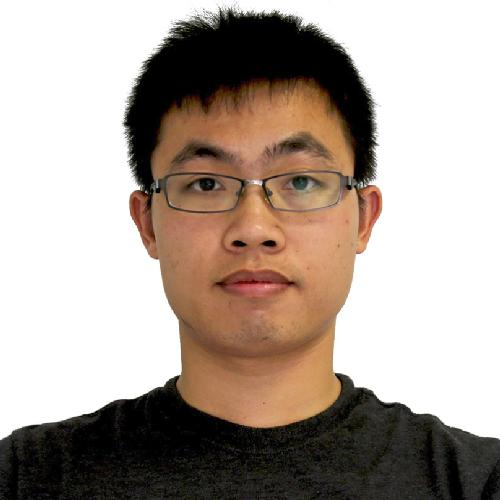}}]{Dihong Gong}
received the Ph.D. degree of computer science from the University of Florida, in 2018. He then joined the Tencent AI Lab as a senior research scientist, with research interest primarily focused on face related technologies, including face detection, recognition, liveness examination, etc.
\end{IEEEbiography}
\vspace{-1em}

\begin{IEEEbiography}[{\includegraphics[width=1in,height=1.25in,clip,keepaspectratio]{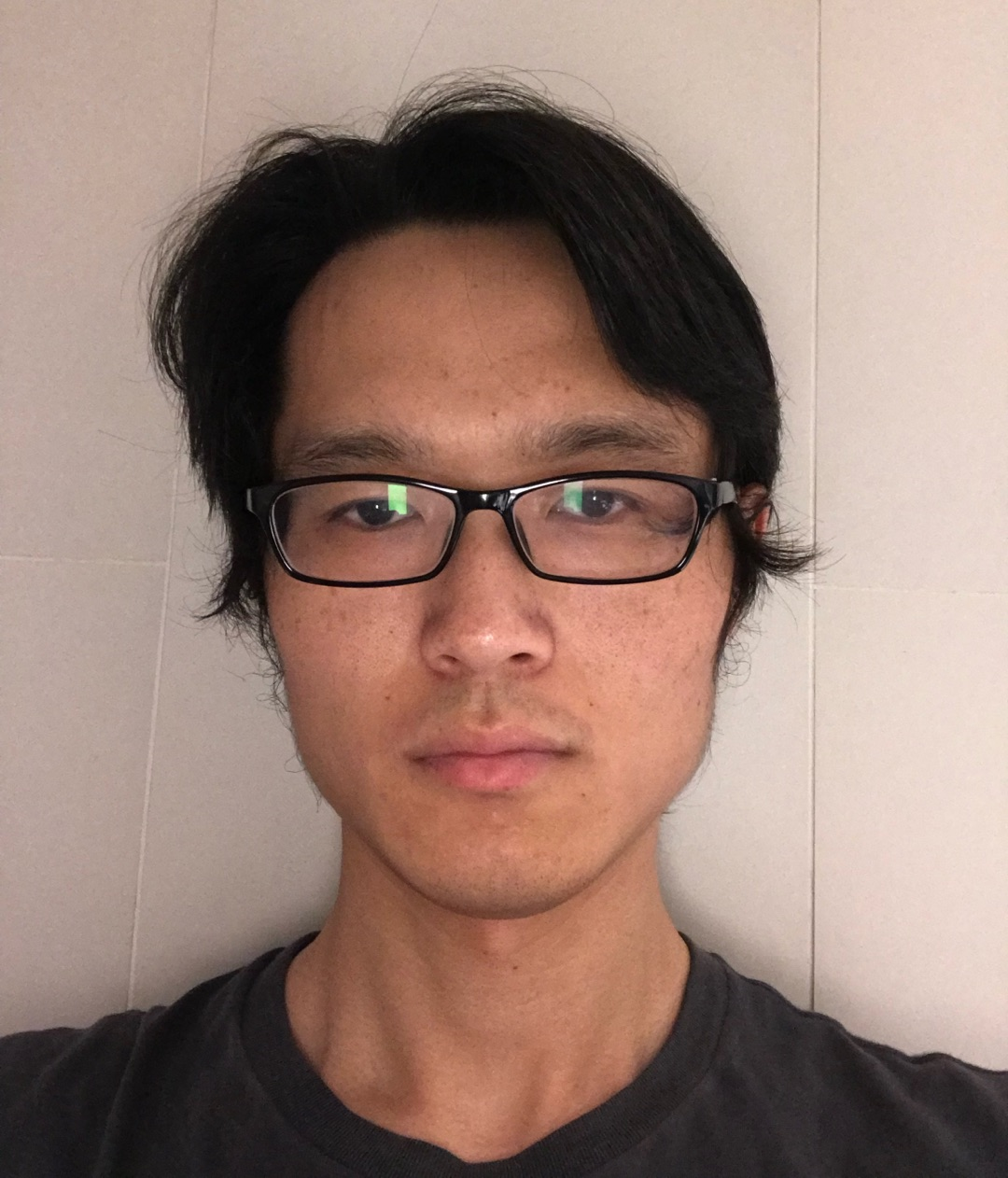}}]{Hao Wang}
is currently a senior researcher with Tencent AI Lab. He received the M.Eng. degree from the University of Science and Technology of China in 2016. His research interests include computer vision, deep learning, and face recognition.
\end{IEEEbiography}
\vspace{-1em}

\begin{IEEEbiography}[{\includegraphics[width=1in,height=1.25in,clip,keepaspectratio]{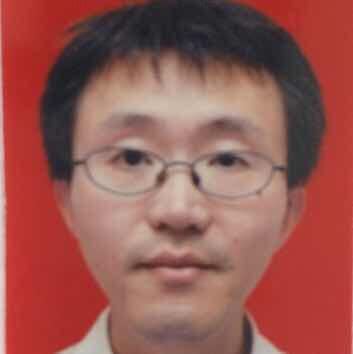}}]{Zhifeng Li}
(M'06-SM'11) is currently a top-tier principal researcher with Tencent. He received the Ph. D. degree from the Chinese University of Hong Kong in 2006. After that, He was a postdoctoral fellow at the Chinese University of Hong Kong and Michigan State University for several years. Before joining Tencent, he was a full professor with the Shenzhen Institutes of Advanced Technology, Chinese Academy of Sciences. His research interests include deep learning, computer vision and pattern recognition, and face detection and recognition. He is currently serving on the Editorial Boards of Neurocomputing and IEEE Transactions on Circuits and Systems for Video Technology. He is a fellow of British Computer Society (FBCS). 
\end{IEEEbiography}
\vspace{-1em}

\begin{IEEEbiography}[{\includegraphics[width=1in,height=1.25in,clip,keepaspectratio]{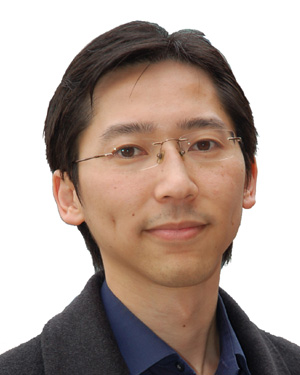}}]{Kwan-Yee K. Wong}
received the BEng degree (Hons.) in computer engineering from The Chinese University of Hong Kong, in 1998, and the MPhil and PhD degrees in computer vision (information engineering) from the University of Cambridge, in 2000 and 2001, respectively. Since 2001, he has been with the Department of Computer Science at The University of Hong Kong, where he is currently an associate professor. His research interests are in computer vision and machine intelligence. He is currently an editorial board member of International Journal of Computer Vision (IJCV).
\end{IEEEbiography}

\end{document}